\title{MSc Project}
\bigskip
\author{Student: Mr. Andrew Pickin, ap0736\\
Supervisor: Dr. Neill Campbell\\
Marker 1: Dr. Walterio Mayol-Cuevas\\
Marker 2: Professor Majid Mirmehdi\\
Project Type (I/II/III): 25/75/0	}
\medskip
\date{\today}

\documentclass[11pt]{report}
\usepackage{natbib}
\usepackage{graphicx} 
\usepackage[table]{xcolor}
\usepackage{rotating}
\usepackage[total={5.4in,7.8in}]{geometry}
\usepackage{amsmath}
\usepackage{amssymb}
\usepackage{amsthm}
\usepackage{url}
\usepackage{lipsum}% just to automatically generate text
\usepackage{listings}

\makeatletter
\newcommand\ackname{Acknowledgements}
\if@titlepage
  \newenvironment{acknowledgements}{%
      \titlepage
      \null\vfil
      \@beginparpenalty\@lowpenalty
      \begin{center}%
        \bfseries \ackname
        \@endparpenalty\@M
      \end{center}}%
     {\par\vfil\null\endtitlepage}
\else
  \newenvironment{acknowledgements}{%
      \if@twocolumn
        \section*{\abstractname}%
      \else
        \small
        \begin{center}%
          {\bfseries \ackname\vspace{-.5em}\vspace{\z@}}%
        \end{center}%
        \quotation
      \fi}
      {\if@twocolumn\else\endquotation\fi}
\fi
\makeatother

\makeatletter
\newcommand\sumname{Executive Summary}
\if@titlepage
  \newenvironment{summary}{%
      \titlepage
      \null\vfil
      \@beginparpenalty\@lowpenalty
      \begin{center}%
        \bfseries \sumname
        \@endparpenalty\@M
      \end{center}}%
     {\par\vfil\null\endtitlepage}
\else
  
\fi
\makeatother

\begin{document}
\pagenumbering{gobble}
\begin{center}
{\bf Executive Summary}
\end{center}
This project is about facial asymmetry, its connection to emotional expression, and methods of measuring facial asymmetry in videos of faces. The research was motivated by two factors: firstly, there was a real opportunity to develop a novel measure of asymmetry that required minimal human involvement and that improved on earlier measures in the literature; and secondly, the study of the relationship between facial asymmetry and emotional expression is both interesting in its own right, and important because it can inform neuropsychological theory and answer open questions concerning emotional processing in the brain. The two aims of the research were: first, to develop an automatic frame-by-frame measure of facial asymmetry in videos of faces that improved on previous measures; and second, to use the measure to analyse the relationship between facial asymmetry and emotional expression, and connect our findings with previous research of the relationship. The project is best described as 80\% investigatory and 20\% software development. Since submitting the research review, the main accomplishments have been:

\begin{itemize}\addtolength{\itemsep}{-0.3\baselineskip}
\item the original research review (which forms part of Chapter 2) has been expanded with greater detail on Active Shape Models (Section~\ref{sec:asm}) and Active Appearance Models (Section~\ref{sec:aam}) and a new section (\ref{sec:procrustes}) on Procrustes analysis. (However, it should be noted that the time spent on Chapter 2 was no greater than 25\% of the total time spent on the project.)
\item two original measures of asymmetry were devised and implemented using an Active Appearance Model library and OpenCV (Sections~\ref{sec:measureone} and~\ref{sec:measuretwo}). One of these measures achieved our first aim; i.e., it was automatic and improved on previous measures in the literature.
\item a measure of left-sided and right-sided facial movement (relative to a neutral face) was devised and implemented (Section~\ref{sec:lrmeasure}) and used as a novel measurement of the strength of a subject's emotional expression. The measure was also used to investigate whether one side of the face moved more than the other during the expression of positive emotions (Section~\ref{sec:lrmov}). 
\item the relationship between strength of emotional expression and degree of asymmetry was plotted for subjects expressing laughter and happiness (Figures~\ref{fig:happyasymmetry} and~\ref{fig:edizhappy}). The results gave rise to a hypothesis for future research: that magnitude of facial asymmetry increases with strength of emotional expression, with the increases larger for strong expressions (such as laughter).
\item a thorough critical evaluation of the project was conducted (Section~\ref{sec:evaluation}) and three interesting and fruitful directions for future research were identified (Section~\ref{sec:furtherwork}).
\end{itemize}

\begin{acknowledgements}
I would like to thank
\begin{itemize}

\item Dr. Neill Campbell for his enthusiasm and guidance throughout the duration of the project
\item Mr. Alexander Davies for the time he kindly donated to discuss and demonstrate issues relating to Active Appearance Models
\item Professor Majid Mirmehdi and Dr. Walterio Mayol-Cuevas for their questions, interest and suggestions at the poster and demonstration event 
\item My family and friends for their unconditional support
\end{itemize}
 
\end{acknowledgements}
%\newpage

\tableofcontents
\clearpage
\pagenumbering{arabic}
\chapter{Introduction}

This project is about facial asymmetry, its connection to emotional expression, and methods of measuring facial asymmetry in videos of faces. Psychologists have long been interested in facial asymmetry and its connections to - amongst other things - attractiveness, health and personality type. Distinctly, neuropsychologists have looked at the asymmetry of a moving and expressive face, in the hope that it can teach them about emotional processing in the brain (if one side of the face moves more during expression, then this might be because one side of the brain is more active during expression). Whilst measuring the asymmetry of static and neutral faces is important and interesting, our focus is on measuring the asymmetry of moving and expressive faces.

Before the last decade, measurements of asymmetry were either performed directly by humans (either by perceiving asymmetry by eye or by measuring asymmetry by hand) or by electromyography (detecting electrical activity in the facial muscles, due to their movement). Measurements by hand or eye are limited because they are time-consuming. Measurements by electromyography are limited because they measure changes in electrical activity rather than changes in the appearance of the face, and the two need not be perfectly correlated. Techniques from the field of computer vision provide the opportunity for a less labour-intensive measure of asymmetry, and the collection of larger sets of data. In the last few years there have been some studies on asymmetry and emotional expression that have used techniques drawn from computer vision, but they have been limited in number. We believe that there is a real opportunity to develop a novel measure of asymmetry that requires minimal human involvement and that improves on earlier measures. Furthermore, recent studies have tended to focus on the direction of asymmetry during emotional expression (whether one side moves more than the other) rather than on the magnitude of asymmetry. We believe that interesting results can be obtained by focussing on the magnitude of asymmetry and its relation to the type (e.g. happy or sad) and strength of emotional expression. It is these two beliefs that inform the aims of our research. 

\section{Aims of research}

There are two aims of our research:

\begin{itemize}
\item to develop an automatic frame-by-frame measure of facial asymmetry in videos of faces that improves on previous measures 
\item to use the measure to analyse the relationship between facial asymmetry and emotional expression, and connect our findings with previous research of the relationship

\end{itemize}

\section{Structure of the report}

Excluding this introductory chapter, the report is divided into four chapters. Chapter~\ref{sec:chaptwo} provides the background and context for our research and presents the theory that we will rely on to build our measure of asymmetry. The chapter's first section (Section~\ref{sec:litrev1}) is concerned with demonstrating why facial asymmetry is important and with reporting the results of previous research into the connections between asymmetry and attractiveness, health, personality type and emotional expression. The second section (Section~\ref{sec:litrev2}) concentrates on more recent studies that looked at the relationship between asymmetry and emotional expression. The third and final section (Section~\ref{sec:litrev3}) builds a research toolbox by presenting the theory that will be drawn on in the following chapters.

Chapter~\ref{sec:chapterthree} is directed towards the first aim of our research and describes in detail the design and development of two measures of asymmetry. Although the aim of the project is to develop a single measure, after developing the first measure, a second was developed and compared to the first. Even though one of the measures was ultimately discarded, its development helped us to realise some of the strengths of the measure that was preferred. We end the chapter by discussing the limitations of the preferred measure. 

Chapter~\ref{sec:chapfour} is directed towards the second aim of our research and uses the measure developed in the preceding chapter to analyse two videos of subjects displaying happiness and laughter. Data that were collected were analysed in several ways. Section~\ref{sec:analysis} charts asymmetry by frame number for the first subject and allows us to identify frames of elevated asymmetry. Section~\ref{sec:lrmov} charts movement of the left-side of the face and movement of the right-side of the face by frame number (for the first subject) and allows us to establish if one side moves more than the other. In Section~\ref{sec:asymov} we chart strength of emotional expression against magnitude of asymmetry to see if we can discover a relationship. We end the chapter by repeating the experiments for a second subject (Section~\ref{sec:secondsub}) to see if our results are replicated.

Chapter~\ref{sec:conclusion} is the conclusion. We review the report and project and discuss the strengths and weaknesses of the work undertaken. We refer back to the project's aims to see if they have been accomplished. Possible improvements to the work are discussed and future avenues of research are suggested.

\chapter{Background, context and relevant previous research}\label{sec:chaptwo}
\section{About this chapter}
In this chapter we set the scene for the project by surveying previous research from relevant areas. The chapter is divided into three sections. In Section~\ref{sec:litrev1} we examine previous research by psychologists and neuropsychologists on facial asymmetry that helps to show why we should be interested in facial asymmetry at all. Traditionally, facial asymmetry has been of interest to two groups of researchers. Psychologists have been interested in the asymmetry of neutral (i.e. emotionless) faces and have connected symmetry to attractiveness, health and personality type. Neuropsychologists have been particularly interested in looking at how the level of facial asymmetry changes during emotional expression, because this connects with understanding if emotional processing in the brain is focussed on one particular side of the brain. Our project is more concerned with the latter, but a proper survey on facial asymmetry should not neglect the former, and so we discuss both. 

In Section~\ref{sec:litrev2} we turn our attention to more recent attempts to study facial asymmetry - and how it changes during emotional expression - that use techniques from the field of computer vision. We find that these attempts have been limited in number and in scope. Sections~\ref{sec:litrev1} and~\ref{sec:litrev2} jointly motivate our project; together they show why we should be interested in facial asymmetry and emotional expression and why there is an opportunity to improve on previous measurement techniques.

In the third and final section (Section~\ref{sec:litrev3}) we build the research toolbox that will be used in our approach to measuring asymmetry. In particular, we introduce and describe the theory underlying Active Shape Models and Active Appearance Models - which can be used for modelling faces - and we prove an important theorem from Procrustes Analysis that we will use in Chapter 3.

\section{Psychology, neuropsychology and asymmetry}\label{sec:litrev1}
\subsection{Three kinds of asymmetry}
\begin{figure}[htp]
\centering
\includegraphics[scale=1.6]{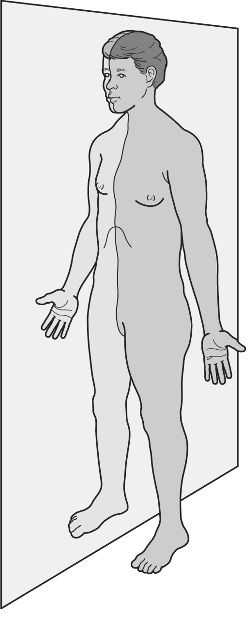}
\caption{The midsaggital plane is the vertical plane through the mid-line of the body, dividing it into two approximately symmetric halves.}\label{fig:midsagg}
\end{figure}

Although the human body exhibits a degree of symmetry across the midsaggital 
plane (Figure~\ref{fig:midsagg}), this symmetry is imperfect, both internally and externally. 
Psychologists have classified the asymmetry into three kinds \citep{gfe94,vvl}. The first kind is \emph{Directional Asymmetry}, which
refers to left-right asymmetry that is characteristic of a population. Some 
directional asymmetries are clear and established for humans: for example, the 
heart is slightly offset to the left-side of the body. Other directional 
asymmetries remain as hypotheses. For example, some studies have found that the 
larger side of the human face is typically the left \citep{vh75}; 
whereas others have found that it is typically the right side \citep{srpk04}. It has been suggested that these differences in 
results may be due to differences in gender or age of the subjects of the 
relevant studies \citep{eoestl}.

The second kind of asymmetry is known as \emph{Antisymmetry} \citep{k01}. Some traits in a population may be naturally dominant on one side of the plane of symmetry (and thus asymmetric); but, in contrast to traits displaying directional asymmetry, the side of dominance is unpredictable. All that is known is that the trait is typically asymmetric in the population.

\begin{figure}[!h]
\centering
\includegraphics[scale=0.36]{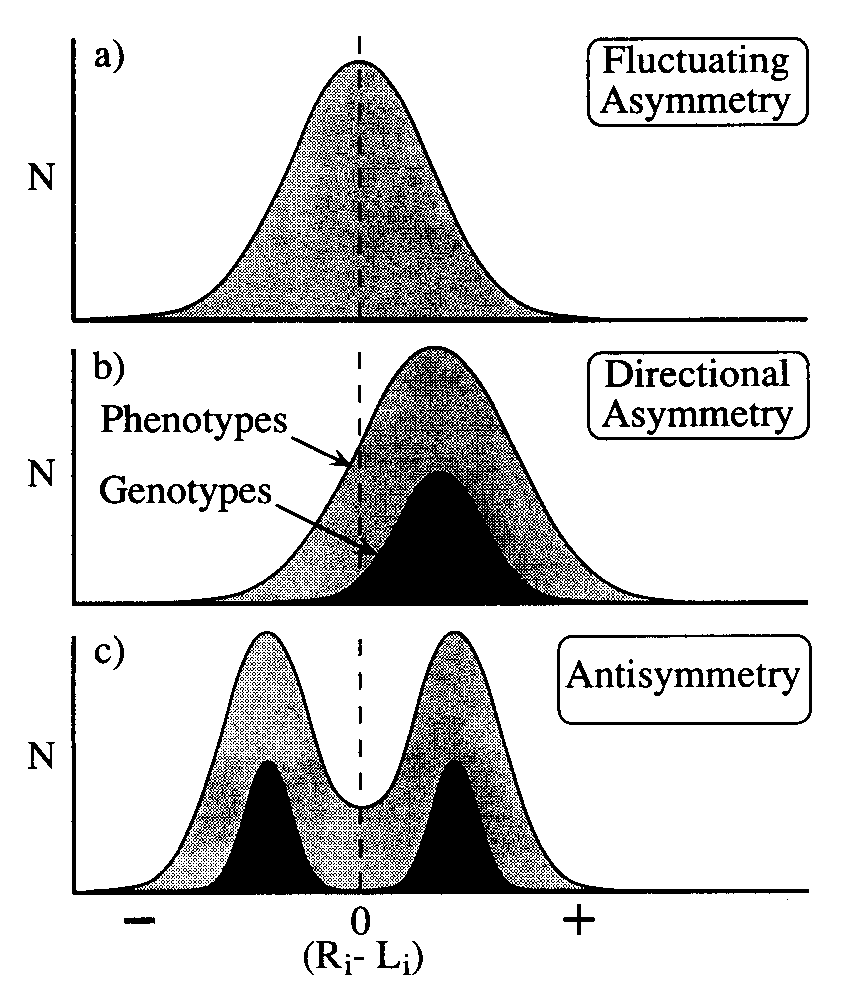}
\caption{Traits exhibiting fluctuating asymmetry have a mean asymmetry of zero, with a normal distribution. Variations in individuals from this mean symmetry is non-genetic. Traits exhibiting directional asymmetry show a skewed normal distribution with mean asymmetry. They are partly genetically determined (i.e., dependent on genotypes). Traits exhibiting antisymmetry show either right-sided asymmetry or left-sided asymmetry, but the direction of asymmetry is unpredictable. (Image taken from \cite{ps92})}\label{fig:charts}
\end{figure}

The final kind of asymmetry is known as \emph{Fluctuating Asymmetry} (FA); the term was first used almost 80 years ago \citep{l32}. In contrast to directional asymmetry
and antisymmetry, fluctuating asymmetry is a property of individuals rather than a property of populations. A trait in an individual is said to display fluctuating 
asymmetry if it is asymmetric in the individual but typically symmetric in the population. This type of asymmetry is known as \emph{fluctuating} because the 
direction of asymmetry appears to be random \citep{k01}. It has been shown that FA is partly heritable, insofar as the magnitude of asymmetry is heritable; but the
direction of asymmetry is not genetically determined. (Figure~\ref{fig:charts} illustrates the distinct distributions associated with the three kinds of asymmetry.)

FA is taken to be a measure of \emph{Developmental Instability} \citep{vvl,z81}. Developmental instability refers to an organism's inability to protect its development against small and random changes to its environment \citep{lds}. Research has shown that developmental instability is negatively correlated with fitness components such as longevity and survival \citep{m97}. Furthermore, reviews of the relationship between sexual selection and developmental instability have indicated mate preference for individuals displaying lower levels of FA \citep{m93,t92,wt94}.

\subsection{Symmetry and attractiveness}
Numerous studies have investigated the relationship between human symmetry and attractiveness to potential mates. Some have considered symmetry of the body in general 
\citep{hm01,rg99} but most have focussed on symmetry of the face in particular, and this will be the focus of our project. Earlier studies found that facial symmetry was 
not positively related to facial attractiveness; in fact, in some studies, a negative relation was found - slightly asymmetric faces were preferred to perfectly symmetric 
ones \citep{k96,lrm94,sc95}. However, from more recent studies, there is a growing body of evidence that facial symmetry is positively correlated with ratings of 
attractiveness for both Western cultures \citep{pbplre99,rrs99} and non-Western cultures \citep{lam07,ryclma01}. Some authors have suggested that symmetric faces are more 
attractive in virtue of their ``averageness'' (separate research has shown that averaging human faces across a population tends to increase attractiveness
\citep{rt96}). However, \citep{rsb99} have used regression analyses to show that symmetry contributes to attractiveness even when the effects of
averageness are partialed out, suggesting that symmetry is independently attractive.

\begin{figure}[!h]
\centering
\includegraphics[scale=0.67]{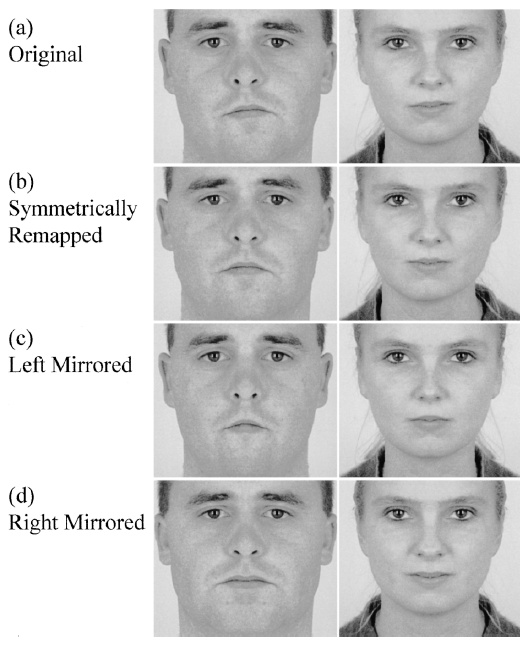}
\caption{The image illustrates the different ways of creating symmetric faces. (a) is the unaltered face; (c) and (d) are obtained by reflecting the right and left hemifaces respectively. (b) is obtained by blending the left mirrored image with the right mirrored image. (Image taken from \cite{pbplre99}).}\label{fig:faces}
\end{figure}

\citep{r06} suggests that the reason that earlier studies found that facial symmetry was not positively correlated with attractiveness is that they relied on a flawed 
methodology. Namely, by simply reflecting one hemiface to create a perfectly symmetric face, certain features of the symmetric face were abnormally large or abnormally 
small. The approach of the later studies was to create two perfectly symmetric faces - one from each hemiface - and then blend the two faces together to create a perfectly 
symmetric face with the abnormalities smoothed out \citep{pbplre99,rrs99,rsb99}. This symmetric face turned out to be more attractive than its asymmetric parents.
Figure~\ref{fig:faces} provides an example of the results of the two approaches. In (c) and (d), symmetric faces are obtained by reflecting the right and left hemifaces respectively. 
This results in unnatural looking features; for example, the man's mouth in (c) looks unnaturally small, whereas his mouth in (d) looks unnaturally large. (b) is the result 
of blending the left mirrored image with the right mirrored face, and is the approach taken by later studies, such as Perrett and colleagues (1999). The large mouth of (d) and 
the small mouth of (c) are blended to create an average and symmetric mouth. 

A recent paper has shown that symmetric faces are not only rated as more attractive, but that symmetry influences the real-world selection of sexual partners
\citep{brwpl11}. As discussed earlier, fluctuating asymmetry is taken to be a measure of developmental instability, and so body symmetry (and facial symmetry) may be 
an indicator of genetic quality. One explanation then for the connections between facial symmetry, attractiveness and selection of sexual partners is that using 
symmetry to guide selection has adaptive value \citep{fp02,srpk04,tg93}. If this explanation is correct, we would expect fluctuating asymmetry in particular to show a 
negative correlation with attractiveness, rather than directional asymmetry or antisymmetry \citep{r06}. Some studies on symmetry and attractiveness have attempted to isolate
fluctuating asymmetry. Simmons and colleagues (2004) found that directional asymmetry affected neither symmetry nor attractiveness judgements, whereas fluctuating asymmetry
affected both. Their conclusion was that humans pick up on fluctuating asymmetry in faces and use it as an indicator of lower genetic quality.

\subsection{Symmetry and other connections}
\subsubsection{Symmetry and health}
Several studies have shown that symmetric faces are perceived as being healthier than less symmetric faces \citep{fnmg06,jlptbp01,rzckhm01}. However, interestingly, a 
link between symmetry and actual health has not yet been established \citep{r06,rzckhm01}. \citep{hm01} found a positive association between body mass index and facial 
asymmetry for females, but this did not extend to an association between reported health problems and facial asymmetry. In response, \citep{srpk04} have suggested that the
cost of developmental instability may be reduced longevity or fecundity rather than poor health. They point to a study by \citep{ha03} that has identified a positive 
relationship between facial attractiveness and longevity, but acknowledge that further research is needed to better understand how asymmetry influences fitness across 
a broad set of measures.

\subsubsection{Symmetry and personality type}

Although not as heavily researched as the relationship between symmetry and attractiveness, there has been some recent interest in connections between facial symmetry 
and perceived and actual personality types.  For example, \citep{ne03} investigated the relationship between facial symmetry and perceived personality across five 
dimensions of personality (known as the ``big-five'' factors � \citep{d90}): neuroticism, extraversion, openness, agreeableness and conscientiousness. Based on the assumption 
that symmetry is used as an indicator of genetic quality, they hypothesised that more symmetric faces would be more likely to be perceived as possessing the desirable 
personality traits - extraversion, openness, agreeableness and conscientiousness - while less symmetric faces would be more likely to be perceived as possessing the 
undesirable trait - neuroticism. Their results - albeit from a small sample size - partially confirmed their hypothesis; they found that asymmetric faces were rated as 
significantly more neurotic, less agreeable and less conscientious. \citep{fnmg05} replicated the research of Noor and Evans but for actual personality rather than 
perceived personality. They found a positive association between facial symmetry and actual extraversion, but - unexpectedly - a negative association between symmetry 
and actual openness. They suggested that further research was necessary to confirm or disconfirm these links. A similar experiment involving a larger data sample of 
participants also found a positive association between symmetry and extraversion \citep{ppb07}.

The causal connection between facial symmetry and extraversion and other personality traits is as yet neither confirmed nor understood. We have seen that symmetry is 
linked to developmental stability and genetic quality, but it is not clear whether it should be linked to actual personality traits. One possibility is that hormones 
such as testosterone and oestrogen affect both facial symmetry and personality jointly, resulting in certain associations. But further research is needed to draw any 
definite conclusions \citep{fnmg06}. 

\subsection{Symmetry and facial expressions}

Our discussion so far has concentrated on the symmetry of motionless faces with neutral expressions. However, there has also been extensive research on the symmetry - and
lack of symmetry - of faces during emotional expression. One of the motivations for this research is that it connects with theory concerning cerebral hemispheric 
specialisation; the human brain is divided into two hemispheres - the left and right hemispheres - and hemispheric specialisation investigates the extent to which cognitive
tasks are separated between these hemispheres. It has been well established that the right hemisphere exerts more control over the muscles of the left hemiface than the 
left hemisphere and vice versa \citep{r84}. Thus, if it is also the case that emotional expressions have higher intensity on one hemiface than the other, then this could
indicate that the respective hemisphere (i.e. right hemisphere for left hemiface and vice versa) is the dominant controller of emotional displays on the face.

The first mention of asymmetry during emotional expression dates back to \citep{d72} who noted that from a sample of four Australian natives who were asked to sneer, 
two displayed the left canine tooth, one displayed the right, and the other displayed no asymmetry. Nearly seventy years later, \citep{ll38} performed the first detailed
study of facial asymmetry during emotional expression and found that although the majority of their subjects did not display significant asymmetry, some of their 
subjects did. They introduced the term ``facedness'' to correspond to the term ``handedness''. A person with left-facedness is a person whose left hemiface is dominant 
(i.e., shows greater movement) during emotional expression.

\begin{figure}[htp]
\centering
\includegraphics[scale=0.8]{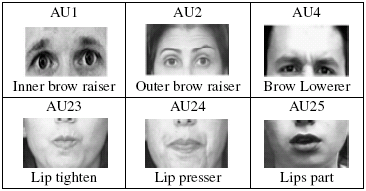}
\caption{Examples of some action units from the Facial Action Coding System (image taken from \cite{z08}).}\label{fig:units}
\end{figure}

\citep{ef69} hypothesised the existence of seven universal (i.e., cross-cultural) emotional facial behaviours to signal seven different emotions: happiness, sadness,
anger, fear, surprise, disgust and interest. During the following decade they worked on coding facial expressions and developed the Facial Action Coding
System \citep{ef78} which works by describing facial expressions as conjunctions of action units (examples of action units are parting the lips or lowering 
the brow - see Figure~\ref{fig:units} for visual examples). 

At the same time, there was a resurgence of interest in the asymmetry of emotional expression and the start of systematic research of the field.
\citep{bkyss98} produced an extensive meta-analysis of research prior to 1998. They considered the results of 82 observations from 35 journal articles to see if areas 
of agreement between the results could be discerned. Of the 82 observations they found that 27 found no significant asymmetry during emotional expression, but that 
48 found left hemiface dominance (i.e., left-facedness) and only 7 found right hemiface dominance. For examples of observations that found left hemiface dominance 
see \citep{bcj,dbtb84}; for an example of observations that found right hemiface dominance see \citep{ss82}; and for examples that found no significant asymmetry 
see \citep{kmm91,sk80}.

\citep{bkyss98} also divided the 82 observations according to various criteria to investigate further hypotheses. The elicitation condition of an emotional expression 
refers to the circumstances that brought it about. Posed (or voluntary) expressions are deliberately produced by the subject, whereas spontaneous (or involuntary) 
expressions are instinctual reactions to a stimulus. Since there have been suggestions in the literature that different neuroanatomical systems are responsible for 
posed and spontaneous expressions (for example, see \cite{bk84}, for a discussion), Borod and colleagues divided the 82 observations according to elicitation condition 
to see if the condition made a difference to the degree or direction of asymmetry in emotional expressions. They found that the literature showed no such 
significant difference \citep{bhk97} with the left hemiface tending to show greater emotional expressivity for both posed and spontaneous expressions.

There are two primary hypotheses about cerebral hemispheric specialisation for emotional processing. One is the right hemisphere hypothesis which holds that the right 
hemisphere is dominant for all emotions irrespective of their valence \citep{bkkma88}. (The valence of an emotion refers to its property of being either positive/pleasant
- emotions such as happiness, interest or pleasant surprise - or negative/unpleasant - emotions such as fear, anger or sadness; for a variation of the positive/negative
distinction see the approach/withdrawal distinction \citep{deyh05}.) The other hypothesis is the valence hypothesis which holds that the right hemisphere is dominant for
negative/unpleasant emotions whereas the left hemisphere is dominant for positive/pleasant emotions \citep{sgwghg82}. Borod and colleagues (1997) divided the 82 
observations according to the valence of the emotional expression observed. They found that negative emotional expressions were more likely to show a left hemiface 
dominance than positive emotions, but that for both negative and positive emotional expressions, a left hemiface dominance was more likely than a right hemiface dominance.
Their conclusion was that there is evidence that emotional valence does influence the direction of asymmetry but that support for the valence hypothesis - which 
predicts right hemiface dominance for positive emotional expressions - is weaker than support for the right hemisphere hypothesis, which predicts left hemiface 
dominance for both positive and negative emotional expressions. They suggested that a further possibility is that positive emotional expressions are mediated by both 
the left and right hemispheres \citep{bhk97}. 

\begin{figure}[htp]
\centering
\includegraphics[scale=0.53]{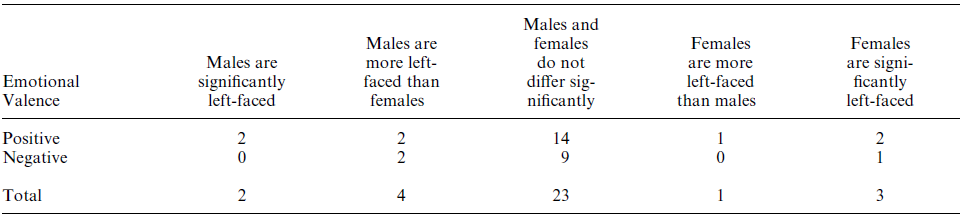}
\caption{The results of 33 observations comparing male facedness to female facedness, separated by emotional valence. Taken from the meta-analysis by \cite{bkyss98}}\label{fig:table1}
\end{figure}

Borod and colleagues (1998) considered two further criteria besides the elicitation condition and emotional valence; namely, gender of the subject and the technique used
to measure the degree of asymmetry. With respect to gender, an earlier study had found that females were more likely than males to show right hemisphere dominance for 
emotional processing \citep{lur80}, and this provided motivation for Borod and colleagues' meta-analysis. They selected the 33 observations where it was possible to 
separate the results by gender and found that of these 33 observations, 23 showed no significant difference between genders; 6 showed that males were more left-faced 
than females; and 4 showed that females were more left-faced than males. They concluded that there was no support for the hypothesis that males were more left-faced 
than females, nor vice versa (see Figure~\ref{fig:table1}).

\begin{figure}[htp]
\centering
\includegraphics[scale=0.53]{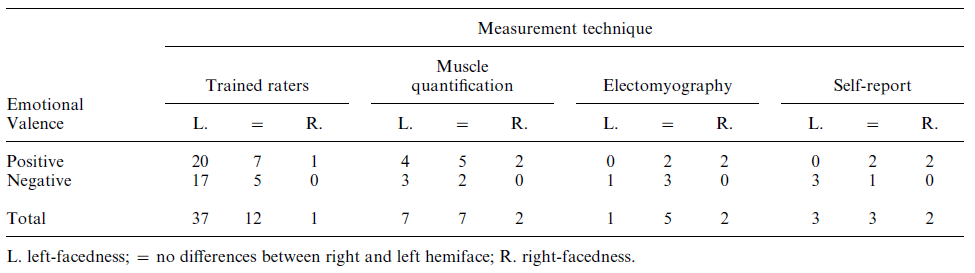}
\caption{The results of 82 observations as a function of measurement technique and emotional valence. Taken from the meta-analysis by \citep{bkyss98}.}\label{fig:table2}
\end{figure}

The final criterion considered by Borod and colleagues was the technique used to measure the degree of asymmetry. From the 82 observations considered in their 
meta-analysis, they identified four categories of measurement techniques. The most popular category was using a group of human raters to decide whether one hemiface 
was more expressive than the other. The second most popular category involved muscle quantification - i.e., making measurements of the movement of muscles away from 
their starting position during emotional expression. The other two categories - both less popular - were electromyography (mostly on the zygomatic muscle) and 
self-report. (Electromyography refers to the technique of measuring muscle activity by recording the electrical activity of those muscles. See \cite{ss82}, for an example.)
Borod and colleagues found that measurement technique made a significant difference to the results of observations with the technique of using human raters showing
a strong tendency to conclude left-facedness, but the techniques of electromyography and self-report tending to find no facial asymmetry. Muscle quantification 
techniques tended to find left-facedness, but not as frequently as 
techniques using human raters (see Figure~\ref{fig:table2}).

\section{Measuring facial asymmetry and computer vision}\label{sec:litrev2}

The four categories of measurement technique identified by \citep{bkyss98} are all limited in one respect or another. Using human raters, self-report or muscle quantification requires human measurement by hand or by eye. Human measurement is subject to human error, but, more importantly, is highly time-consuming and limits the amount of data that can be collected. Measurement using electromyography need not be time-consuming but is limited insofar as it measures changes in electrical activity rather than changes in the appearance of the face, and the two are not perfectly correlated. 

Borod and colleagues were writing in 1998 and computer vision has since developed substantially. In recent years there have some, but not many, applications of techniques from the field to the problem of measuring facial asymmetry during emotional 
expression \citep{d09,necy04,rbbhl00}. In this section we report on two of the more recent ones.

\subsection{3D measurement by Nicholls et al.}
\begin{figure}[htp]
\centering
\includegraphics[scale=0.6]{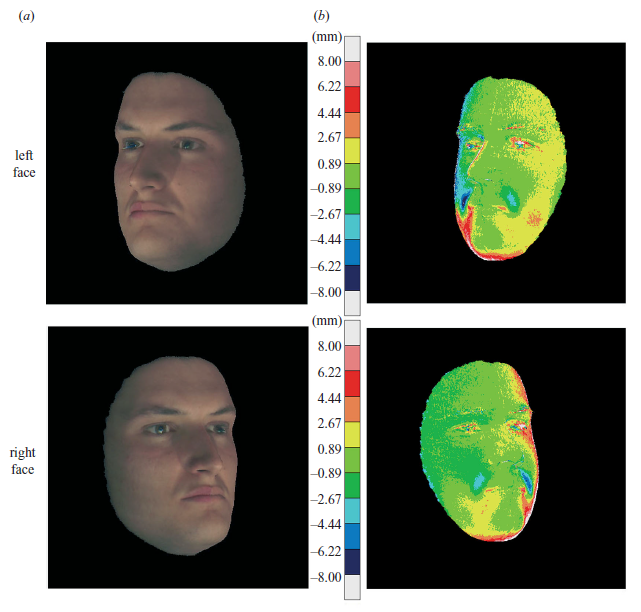}
\caption{(a) shows a posed sad expression rotated by 35 degrees in two directions to display the left side of the face (top) and the right side of the face (bottom). (b) shows colour maps indicating the movement of each part of the face relative to the baseline face (i.e., face with neutral expression - not shown here). The left side of the face (top) shows more movement (indicated by yellow) than the right side of the face (bottom) which is predominantly green. (Image taken from \cite{necy04})}\label{fig:nicholls}
\end{figure}

Nicholls and colleagues used a 3D physiognomic range finder to capture the facial expressions of their subjects. They were then 
able generate 3D images of their subjects which they could use to make 3D measurements of the facial asymmetry under facial expressions. They did this by rotating 
the images to the left and right in turn (by 35 degrees in each direction) and overlaying the image with emotion expressed onto a baseline image with a neutral pose. 
From this they created colour maps showing the amount of movement in different parts of the left and right sides of the face (see Figure~\ref{fig:nicholls} for an example)
and could calculate measures of the overall movement on each side of the face. 
Their experiment only involved measurements for posed happy and sad expressions. They found
that happy expressions involved more facial movement than sad expressions and that both showed a left hemiface dominance, but with the dominance much stronger for sad
expressions than happy expressions. (Notice that this is consistent with the findings of \citep{bkyss98}, as discussed earlier.)

\subsection{Measuring pixel changes by Desai}

\begin{figure}[htp]
\centering
\includegraphics{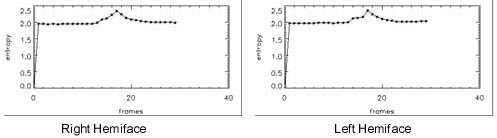}
\caption{Entropy (which is a measure of facial movement between consecutive frames) charted for both the right and left hemifaces, as the face displayed a happy expression. Peak entropy is at frame 17 and is likely to correspond - according to Desai - to the frame in which the facial expression of happiness was strongest. In this case the values of entropy are close for both hemifaces, and so no significant facial asymmetry is detected. (Image taken from \citep{d09})}\label{fig:entropy}
\end{figure}

Desai (2009) videotaped the posed facial expressions of 48 subjects, asking them to produce expressions of happiness, sadness, anger, fear, surprise and disgust. 
The videos were then digitised so that they could be analysed frame by frame and pixel by pixel. Following an earlier study by Richardson and colleagues (2000), Desai 
quantified the degree of facial movement between consecutive frames by computing differences in pixel intensity for all pixels (the frames were 640 x 480 pixels at 256 
levels of grayscale). The entropy was calculated as the sum of pixel intensity differences across all corresponding pixels in the two consecutive frames. Entropy could 
then be charted by frame to identify pairs of consecutive frames where the overall change in pixel intensity was greatest. Working on the assumption that ``changes in 
the surface lighting of the face reflect movement'', entropy was taken as an objective measure of facial movement. By dividing the face into left and right hemifaces, 
and recording entropy scores for each hemiface separately, facial asymmetry could also be measured (see Figure~\ref{fig:entropy} for an example of charting entropy).
Using this approach, Desai found that males displayed left hemiface dominance for all emotional expressions, whereas females displayed less significant left hemiface 
dominance, and even showed right hemiface dominance for expressions of sadness. This latter result runs counter to the findings of Borod and colleagues (1998). Desai's 
explanation for the difference in asymmetry for males and females was that emotional processing may be less lateralised in the female brain than in the male brain, 
as seems to be the case for linguistic processing (e.g., see \cite{sspcsfbfskg95}). 
\section{Building a research toolbox}\label{sec:litrev3}
Both \cite{necy04} and \cite{d09} admit to limitations of their measures. Nicholls and colleagues stated that their measure was only sensitive to movement perpendicular to the surface of the face and Desai assumed that ``changes in the surface lighting of the face reflect movement'' meaning that the measure would be affected by lighting variation. Furthermore, Desai's analysis of videos was limited to short snippets of a few seconds, and Nicholls and colleagues looked only at whether one side of the face moved more than the other, and did not employ a measure of asymmetry in their analysis. 

We now present the theory that will be used to develop an automatic measure of facial asymmetry that improves on earlier measures, and that can be used to analyse the relationship between facial asymmetry and emotional expression in ways neglected by \cite{necy04} and \cite{d09}. 

\subsection{Active Shape Models}\label{sec:asm}
Active Shape Models (ASMs) are statistical models - developed in the 1990s - that have been used in a 
variety of contexts. For example, ASMs have been used in medical imaging to analyse MR images of the brain and locate structures \citep{hctl94} and to identify bones in 
radiographs of hip replacements \citep{krth96}. Other examples of applications are their use in visual speech recognition \citep{ltb96}, person identification from faces \citep{ltc97}, facial 
expression recognition \citep{ltc97}, and in classifying crops in images \citep{pa08}.

\subsubsection{Training}

ASMs were first developed by Cootes and Taylor \citep{ct93}. There are two steps involved in their use; the \emph{training} step and the \emph{fitting} step. The training step requires 
a training set of images of the relevant kind of object. For example, if we are interested in face recognition then we need a set of images of faces. Training sets should
contain a variety of images of the relevant kind of object, so that shape variations can be fully modelled \citep{ctcg95}. For instance, if we are building a training set of human
faces that will be used for person identification from faces with various expressions, presented at different angles, then we should ensure that we include faces of different subjects, with various facial expressions,
presented at different angles, in our training set. Generally a larger training set is preferred (for example, \cite{ltc97}, use 160 images of faces in their training set),
but of most importance is the variety of image within the set. Adequate, simpler models can be built from as few as 10-20 images in the training set. 

\begin{figure}[htp]
\centering
\includegraphics[scale=0.65]{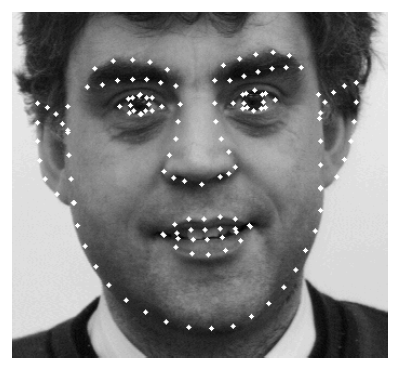}
\caption{A training image of a face annotated for use in an Active Shape Model. (Image taken from \cite{ltc97})}\label{fig:annotated}
\end{figure}

Each image in the training set must be annotated to obtain a set of points which represents the relevant shape (for example, see Figure~\ref{fig:annotated}). 
 Annotation is typically done by
hand, but methods have been developed for automatic annotation \citep{su05}. Since the training images are 2-dimensional, each set of points can be represented by a vector of 2n-dimensions, where n is the number of
points marked in each training image. If each point is represented as $(x_{i}, y_{i})$ then we can write the vector as:
\[\mathbf{x} = (x_{1},...,x_{n},y_{1},...,y_{n})\]
We can think of each set of points as representing a shape. The mean shape is computed by averaging the N vectors (where N is the number of images in the training set) and the mean shape is centred on the origin (by translation) and scaled so that the sum of the squares of its coordinates is 1. Each shape in the training set is then aligned to the mean shape using rotations, translations and scaling. This is typically accomplished using Procrustes analysis (we present Procrustes analysis in more detail in Section~\ref{sec:procrustes}). Alignment in this instance means minimising the sum of the squares of the distances between corresponding points of each shape and the mean shape. If the two shapes are $(x_{1},...,x_{n},y_{1},...,y_{n})$ and $(x'_{1},...,x'_{n},y'_{1},...,y'_{n})$ then we want to minimise D, where:
\[D = \sum\limits_{i=1}^n (x_i - x'_i)^2 + (y_i - y'_i)^2\]
After alignment, we represent the set of N training shapes as a matrix with 2n rows and N columns. Call this matrix $\mathbf{M}$. The matrix represents the range of shapes in the training set. If the $j^{th}$ training shape is represented by the vector $(x_{1j},...,x_{nj},y_{1j},...,y_{nj})$ then the matrix $\mathbf{M}$ has the form:
\[ \mathbf{M} = \left(\begin{array}{ccc}
x_{11}&...&x_{1N}\\
\vdots& &\vdots\\
x_{n1}&...&x_{nN}\\
y_{11}&...&y_{1N}\\
\vdots& &\vdots\\
y_{n1}&...&y_{nN}\end{array} \right)\] 
Principal Components Analysis (PCA) \citep{jw88} is applied to $\mathbf{M}$ to calculate the set of principal components for the set of shapes in the training set. The first step in PCA is to calculate the 2n by 2n covariance matrix for $\mathbf{M}$. The covariance matrix is given by:
\[\mathbf{S} = \frac{1}{N-1}\sum\limits_{i=1}^N (\mathbf{x_i} - \mathbf{\bar{x}})(\mathbf{x_i} - \mathbf{\bar{x}})^T\]
where $\mathbf{x_i} = (x_{1j},...,x_{nj},y_{1j},...,y_{nj})^T$ and $\mathbf{\bar{x}}$ is the mean shape given by
\[\mathbf{\bar{x}} = \frac{1}{N}\sum\limits_{i=1}^N \mathbf{x_i}\]
The second step is to calculate the 2n (eigenvalue, eigenvector) pairs for $\mathbf{S}$. An (eigenvalue, eigenvector) pair for $\mathbf{S}$ is a (scalar, vector) pair, ($\lambda$, $\mathbf{v}$), such that
\[\mathbf{Sv} = \lambda\mathbf{v}\]
For large n, calculating the eigenvalues and eigenvectors is non-trivial. There exist various algorithms and software implementations (in, for example, MATLAB and OpenCV) that perform the calculation. 

The eigenvectors are ordered by the
magnitude of their associated eigenvalues; the eigenvector with the largest eigenvalue is the principal component, and the eigenvector with the second largest eigenvalue 
is the second principal component, and so on. Although there are 2n eigenvectors, it may be that the first t of these (for some t less than 2n) capture most of the shape 
variation in the set of shapes. (Imagine, for example, a set of points in 2-dimensions that are clustered along a straight line; although 2 vectors are needed to capture all of the variation of the points, a single vector along the straight line will capture most of the variation.) The eigenvalue of an eigenvector equals the variance that the eigenvector accounts for; this means that the sum of the eigenvalues is the total variance of the training data. t is chosen so that the first t eigenvectors represent nearly all, but not necessarily all, of the variation. I.e., t is chosen so that:
\[\sum\limits_{i=1}^t \lambda_{i} \geq pV\] 
where p is the proportion of variation that we wish the t eigenvectors to capture (e.g., p = 0.98) and V is the total variance of the training data (i.e., V = $\sum\limits_{i=1}^{2n} \lambda_{i}$)

The t eigenvectors are retained - and are used in the ASM - and the other eigenvectors are discarded. New shapes can be generated by taking the mean shape and adding linear combinations of the t eigenvectors. If $\mathbf{A}$ is the matrix of t eigenvectors, then a new shape, $\mathbf{x}$, generated by the ASM, has the form:
\[\mathbf{x} = \mathbf{\bar{x}} + \mathbf{A}\mathbf{b}\]
where $\mathbf{\bar{x}}$ is the mean shape and $\mathbf{b}$ is a t-dimensional vector containing the shape parameters for the new shape. Training data can be used to put constraints on likely values for $\mathbf{b}$. This is done by recording the range of values of $\mathbf{b}$ from the training data. Assuming that the range of possible values of $\mathbf{b}$ can be modelled by a Gaussian distribution, we can calculate the mean and variance for this distribution from the training data, and use these to set limits for a plausible range of expected values of $\mathbf{b}$.

\begin{figure}[htp]
\centering
\includegraphics[scale=0.65]{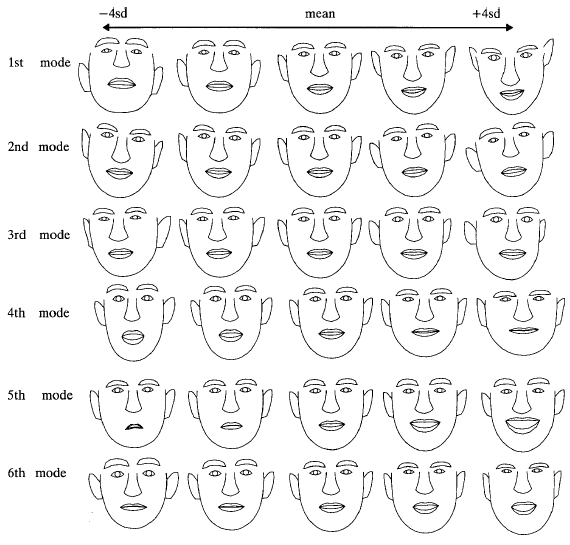}
\caption{The image shows a set of face shapes generated by an Active Shape Model in \citep{ltc97}. In the top row new shapes
are generated by adding multiples of the principal eigenvector (i.e., largest associated eigenvalue) to the mean shape. In the second row new shapes are generated by 
adding multiples of the second principal eigenvector to the mean shape.}\label{fig:modes}
\end{figure}

Figure~\ref{fig:modes} shows a set of face shapes generated by an Active Shape Model. The top row shows shapes generated 
by varying the contribution made by the first eigenvector (i.e. eigenvector with largest associated eigenvalue). This is called
the first mode of variation. The middle face 
is the mean shape (zero eigenvector contribution). The second row shows shapes generated by varying the contribution made by 
the second eigenvector (second mode of variation), and so on. Different eigenvectors produce different behaviours. For example, the first 
mode of variation tilts the face up and down, whereas the fifth changes the facial expression.

\subsubsection{Fitting}

\begin{figure}[htp]
\centering
\includegraphics[scale=0.2]{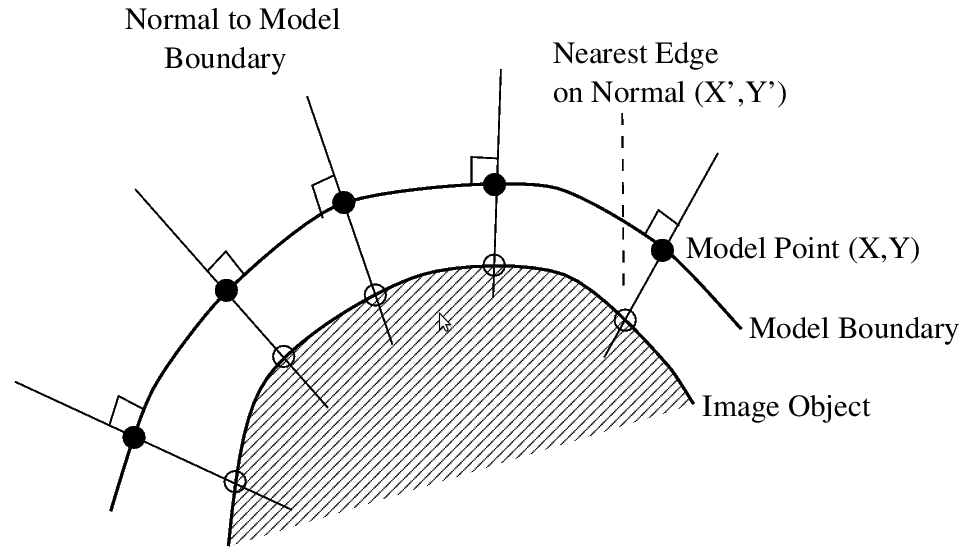}
\caption{The goodness of a shape model's fit to an image is measured by summing the squares of the distances between each model point and the nearest image boundary. (Image taken from \cite{ct012})}\label{fig:boundary}
\end{figure}

Once an ASM has been created from a training set, the model can be used to attempt to fit new images. Suppose we are presented with a new image. The goal of fitting is to find the vector of shape parameters, $\mathbf{b}$, the rotation, $\theta$, the scale factor, s, and the translation vector, $(x_{t}, y_{t})$, such that the shape $\mathbf{x}$ comes closest to fitting the image, where:
\[\mathbf{x} = T_{x_{t}, y_{t}, s, \theta}(\mathbf{\bar{x}} + \mathbf{A}\mathbf{b})\]
(T represents the transformation that translates by $(x_{t}, y_{t})$, scales by s and rotates by $\theta$.)
To make sense of this we need a metric that quantifies the goodness of the shape's fit to the image. Providing that we have a means of detecting edges in the image and defining boundaries, we can use the sum of the squared distances between each point in the shape and the nearest boundary point of the image \citep{ctcg95}. Figure~\ref{fig:boundary} provides an illustration of this.

\begin{figure}[htp]
\centering
\includegraphics[scale=0.25]{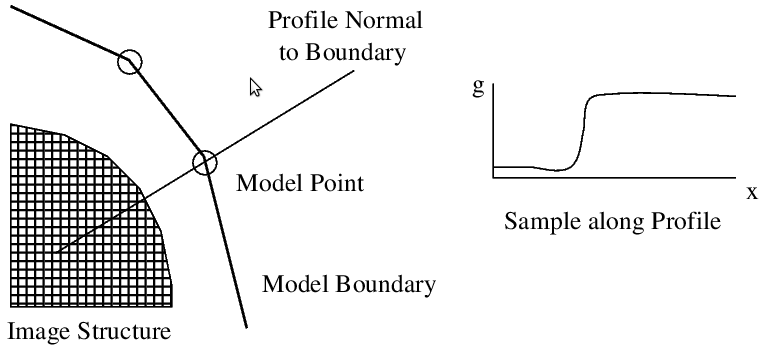}
\caption{Perpendiculars are extended from model points to find the nearest relevant boundary points. Changes in grayscale pixel information are used to detect boundaries. Training information is used to teach the model to correctly identify boundaries by sampling the pixel information that surrounds the landmarks of the training images. (Image taken from \cite{ct012})}\label{fig:boundary2}
\end{figure}

Fitting is performed iteratively. For a new image, the first shape used is the mean shape. If fitting is being applied to a video, then the shape that was used to fit the previous frame is used as the starting shape. Call the candidate model points $\mathbf{x}$. Perpendicular lines from each model point are extended until edges are detected (see Figure~\ref{fig:boundary2}).
The point where the perpendicular intersects the edge is recorded. The collection of these points is called the set of image boundary points (call them $\mathbf{y}$). Procrustes analysis is used to find the $(\theta, (x_{t}, y_{t}), s)$ combination whose associated transformation, when applied to $\mathbf{x}$, minimises the distance between $\mathbf{y}$ and $\mathbf{x}$. The inverse of this transformation is applied to $\mathbf{y}$. The shape parameters, $\mathbf{b}$, are then updated using:
\[\mathbf{b} = \mathbf{A}^{T}(\mathbf{y} - \mathbf{\bar{x}})\]
The constraints on $\mathbf{b}$ that were derived in the training stage are applied to $\mathbf{b}$ (i.e., if $\mathbf{b}$ does not fall within the plausible range of values that was derived from the training data, it is scaled so that it does). The distance between the shape parameters that we started with and the shape parameters that we ended with is measured. If this is above a certain threshold, then the same steps are repeated using the updated parameters. If the distance is below the threshold, then the fitting is completed. (The threshold is determined by the fitting accuracy we desire and the time allowed for each fitting. It is chosen by trial and error.)

\begin{figure}[htp]
\centering
\includegraphics[scale=0.45]{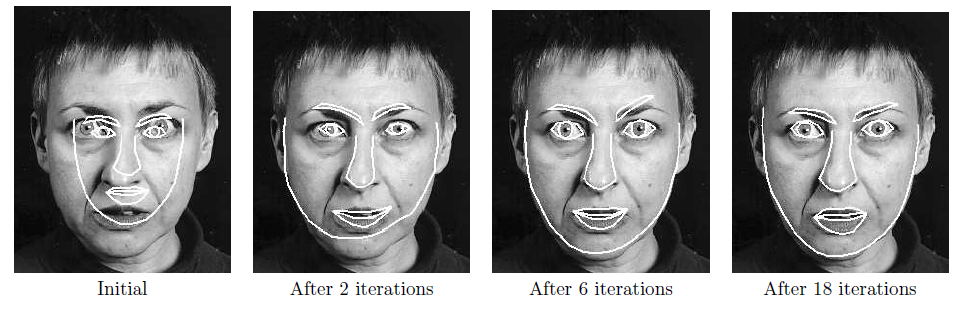}
\caption{Fitting an ASM to an image of a face. (Image taken from \cite{ct01})}\label{fig:fitting}
\end{figure}

Figure~\ref{fig:fitting} illustrates fitting an ASM to an image of a face. In this case, the result can be used to locate 
facial features. Fitting was completed in 18 iterations and in under a second \citep{ct01}. 
As discussed earlier, ASMs have been used to locate structures in MR images
of the brain \citep{hctl94}, and to identify bones in radiographs of hip replacements \citep{krth96}. They can also be used in facial expression identification, by
fitting faces and determining whether the fitted model corresponds to particular expressions in the training set (i.e., whether the model parameters that fit the face are 
close to parameters which represent a particular expression in the training set) \citep{ltc97}.

\subsection{Active Appearance Models}\label{sec:aam}

Active Appearance Models (AAMs) are an extension to ASMs developed by \citep{cet98} which, like ASMs, have had many applications, such as use in medical imaging \citep{bctpt08,ropmca11}, face
tracking and recognition \citep{ect98} and facial expression recognition \citep{Ashraf20091788}.

They extend ASMs by not only modelling shape, but modelling texture as well. This is done by training the model for texture as well as for shape.

\subsubsection{Training}

Training an AAM starts
by training an ASM; i.e., as with ASMs, training images are annotated, and Principal Components Analysis (PCA) is used to compute the modes of variation for the set of training
shapes. Following this, each image in the training set is warped to the mean shape, to effectively factor out the shape information so that the texture information can be sampled independently. The gray-level information from each warped training image
is sampled. Each training image gives rise to a vector, $\mathbf{g_{i}}$,  containing the gray-level information from the warped training image. The vectors are normalised to minimise the effect of global lighting variation across the training images. PCA is then applied to the set of normalised vectors to build a model of the gray-level information. As with PCA on the shape data, we are then able to express texture vectors as linear combinations of the eigenvectors added to the mean vector:
\[\mathbf{g} = \mathbf{\bar{g}} + \mathbf{A'}\mathbf{b}\]
where $\mathbf{\bar{g}}$ is the mean of the normalised vectors, $\mathbf{A'}$ is the matrix of eigenvectors and $\mathbf{b}$ is the vector of texture parameters. 

Because there is likely to be interdependence between shape and texture information, PCA is applied to the combined shape and texture information, to further simplify the model. Each training image has an associated set of shape parameters, $\mathbf{b_{is}}$, and an associated set of texture parameters, $\mathbf{b_{ig}}$. The shape parameters are scaled and then placed in a vector with the texture parameters. This gives rise to a set of vectors, ${\mathbf{b_{i}}}$, where
\[\mathbf{b_{i}} = \left(\begin{array}{c}
\mathbf{W}\mathbf{b_{is}}\\
\mathbf{b_{ig}}\end{array} \right)\] 
and $\mathbf{W}$ is a scaling matrix. (The reason why we apply $\mathbf{W}$ is simply that shape parameters have units of distance whereas texture parameters have units of intensity and the two are therefore not directly comparable. \citep{ct012} suggest defining $\mathbf{W} = r\mathbf{I}$ where $\mathbf{I}$ is the identity matrix and r is the square root of the total intensity variation divided by the total shape variation.)

PCA is applied to the set of vectors, ${\mathbf{b_{i}}}$, and yields a matrix of eigenvectors, $\mathbf{A''}$, such that each  ${\mathbf{b_{i}}}$ can be expressed as:
\[\mathbf{b_{i}} = \mathbf{A''c}\]
$\mathbf{c}$ is the vector of appearance parameters. (Notice that there is no mean element in this formula; the reason is that the shape and texture parameters both have zero mean.) 
Shapes and textures can then be expressed using the appearance parameters. For example, since a texture can be written as:
\[\mathbf{g} = \mathbf{\bar{g}} + \mathbf{A'}\mathbf{b}\]
and since the texture parameters $\mathbf{b}$ can be written as:
\[\mathbf{b} = \mathbf{A''_{g}c}\]
where $\mathbf{A''_{g}}$ denotes the rows of the matrix $\mathbf{A''}$ that relate to texture (i.e., the bottom t rows of the matrix, where t is the number of texture parameters), we have:
\[\mathbf{g} = \mathbf{\bar{g}} + \mathbf{A'}\mathbf{A''_{g}c}\]
Similarly, shapes can be expressed in terms of the appearance parameters, where:
\[\mathbf{x} = \mathbf{\bar{x}} + \mathbf{A}\mathbf{W^{-1}}\mathbf{A''_{s}c}\]
where $\mathbf{A''_{s}}$ denotes the rows of the matrix $\mathbf{A''}$ that relate to shape (and $\mathbf{W}$ is the scaling matrix from above).

\begin{figure}[!h]
\centering
\includegraphics[scale=0.32]{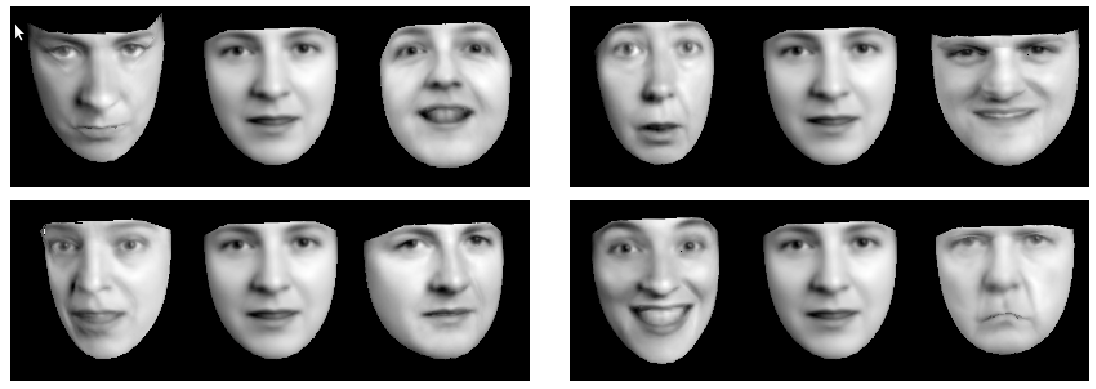}
\caption{The images illustrate 4 of the modes of variation for an AAM trained on 400 images of faces. (Image taken from \cite{ct012})}\label{fig:aamtraining}
\end{figure}

New, complete images (i.e., with shape \emph{and} texture) can be generated by the AAM by using the texture model to generate a mean shape face with texture, and then warping this
face to give it shape by using the shape model. Figure~\ref{fig:aamtraining} is analogous to Figure~\ref{fig:modes} but for an AAM rather than an ASM. Each set of 3 images shows the effect of a mode of variation. The middle image in each set of 3 is the mean image (mean shape and mean texture) and the adjacent images are the effect of the mode of variation.

\subsubsection{Fitting}
As with ASMs, Active Appearance Models can be fitted to images iteratively (see \cite{cet98} for further details). The goodness of fit of a synthesised image generated by the model is measured by taking the square of the norm of $\delta\mathbf{I}$ where 
\[\delta\mathbf{I} = \mathbf{I_i} - \mathbf{I_m}\]
and $\mathbf{I_i}$ is the vector of gray-level values for the target image and $\mathbf{I_m}$ is the vector of gray-level values for the synthesised image. 

Given a new target image, fitting is performed as follows. The starting image is either the mean image (i.e., mean shape and mean texture) or, if we are fitting a video, the fitting from the previous frame is used as the starting image. The error vector $\mathbf{I_{i}} - \mathbf{I_{m}}$ and error value $|\delta\mathbf{I}|^2$ are calculated. A suggested adjustment to the appearance parameters is then computed. The suggested adjustment is derived from information collected from the training data. It turns out that the suggested adjustment, $\delta\mathbf{c}$, (remember that $\mathbf{c}$ denotes the appearance parameters) can be linked to the error vector by the relationship:
\[\delta\mathbf{c} = \mathbf{A}\delta\mathbf{I}\]
where $\mathbf{A}$ is a matrix that can be found by using the training data (where we know the correct appearance parameters) to test different values of $\delta\mathbf{c}$ (by perturbing the appearance parameters by small amounts) and observing the effect on $\delta\mathbf{I}$.
The appearance parameters are updated according to the suggested adjustment multiplied by a scalar, $k$, where $k$ = 1. That is, the appearance parameters become $\mathbf{c} - k\delta\mathbf{c}$. The error value is then recalculated. If it is less than the original error, then we return to the start of iterative process and recalculate the error vector. If it is greater than the original error, then we try the adjustment with k = 0.5. If the error is now less than the original error, then we return to the start of the iterative process. If not, we try k = 0.25. If at this point we find that the error is not improved then convergence is deemed to have occurred and the appearance parameters are accepted. 

\begin{figure}[!h]
\centering
\includegraphics[scale=0.30]{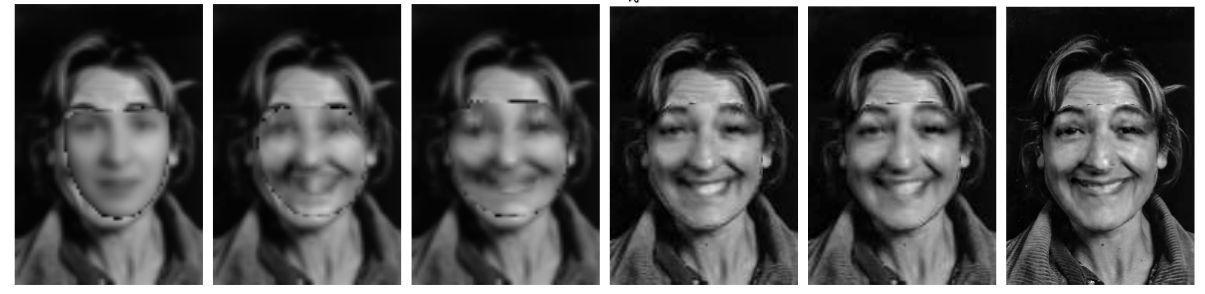}
\caption{The images show an AAM being fitted to a face. Convergence occurs after more than 20 iterations. (Image taken from \cite{cet98})}\label{fig:aamfitting}
\end{figure}

Figure~\ref{fig:aamfitting} shows an AAM being fitted to a face. The left-most image is the starting shape and texture, which clearly does not fit well. The next image shows the synthesised image after 2 iterations; the next after 8 iterations; then 14 iterations and then 20 iterations. The right-most shows the image when convergence has occurred and fitting is complete.

\subsection{Procrustes Analysis}\label{sec:procrustes}

We saw in Section~\ref{sec:asm} that Procrustes analysis is used to align training shapes with the mean training shape in one of the first steps of training an ASM. We also saw that Procrustes analysis is used again during fitting an ASM to an image. Procrustes analysis will also play a crucial role when we develop two measures of asymmetry in the next chapter. In this section we prove a theorem that will be used in the next chapter (and that is used in implementations of ASMs).

Procrustes analysis is a set of techniques in statistical shape analysis that is concerned with shape comparison and solutions to various ``Procrustes problems'' (the term ``Procrustes analysis'' dates back to \cite{hc62}). Of particular relevance to our work is the ``Orthogonal Procrustes Problem'' which was solved by \cite{s66}. Suppose that we have two sets of 2-dimensional points, A and B, with each set containing N points (note that the problem and solution can be generalised to n-dimensions, but we shall only consider 2 dimensions for simplicity). Suppose further that each set of points is normalised so that the sum of the squares of their coordinates is 1 and that each set of points is centred on the origin (i.e., $\sum\limits_{i=1}^N x_{ai} = \sum\limits_{i=1}^N x_{bi} = \sum\limits_{i=1}^N y_{ai} = \sum\limits_{i=1}^N y_{bi} = 0$. A and B can be represented as matrices:
\[\mathbf{A} = \left( \begin{array}{cc}
x_{a1} & y_{a1}\\
x_{a2} & y_{a2}\\
\vdots & \vdots\\
x_{aN} & y_{aN}
\end{array} \right)
\hspace{12pt}\mathbf{B} = \left( \begin{array}{cc}
x_{b1} & y_{b1}\\
x_{b2} & y_{b2}\\
\vdots & \vdots\\
x_{bN} & y_{bN}
\end{array} \right)\]
The Orthogonal Procrustes Problem is to find the orthogonal 2 by 2 matrix, $\mathbf{R}$, that, when applied to $\mathbf{A}$, minimises the Procrustes distance to $\mathbf{B}$. The Procrustes distance between $\mathbf{A}$ and $\mathbf{B}$ is defined as:
\[D = (\sum\limits_{i=1}^N (x_{ai} - x_{bi})^2 + (y_{ai} - y_{bi})^2)^{\frac{1}{2}}\]
In geometric terms, this amounts to finding the rotation and/or reflection that - when applied to the first set of points - minimises the Procrustes distance between the 2 sets of points. 
\newline
\\
\emph{Theorem.}
The orthogonal matrix $\mathbf{R}$ that minimises the Procrustes distance is given by:
\[\mathbf{R} = \mathbf{U}\mathbf{V}^T\]
where
\[\mathbf{B}^T\mathbf{A} = \mathbf{V}\mathbf{W}\mathbf{U}^T\]
is the singular value decomposition of $\mathbf{B}^T\mathbf{A}$ with $\mathbf{U}$ and $\mathbf{V}$ orthogonal matrices and $\mathbf{W}$ a diagonal matrix.

\begin{proof}
The proof is based on the proof in \citep{dm98} with some details filled in.
\newline
\\
We want to minimise $||\mathbf{B} - \mathbf{AR}||$ where $\mathbf{R}$ is orthogonal (i.e., $\mathbf{R}\mathbf{R}^T = \mathbf{R}^T\mathbf{R} = \mathbf{I}$). This is the same as minimising $trace((\mathbf{B} - \mathbf{AR})^T(\mathbf{B} - \mathbf{AR}))$.
The equivalence follows from the definition of Procrustes distance, and the definition of the trace of a matrix as the sum of its diagonal elements. 
\newline
\\
Now, we have the following:
\begin{align*}&trace((\mathbf{B} - \mathbf{AR})^T(\mathbf{B} - \mathbf{AR})) \\ =\hspace{5pt}
&trace((\mathbf{B}^T - \mathbf{R}^T\mathbf{A}^T)(\mathbf{B} - \mathbf{AR})) \\
=\hspace{5pt} &trace(\mathbf{B}^T\mathbf{B}) - trace(\mathbf{R}^T\mathbf{A}^T\mathbf{B}+\mathbf{B}^T\mathbf{AR}) + trace( \mathbf{R}^T\mathbf{A}^T\mathbf{AR}) \\ 
=\hspace{5pt}&trace(\mathbf{B}^T\mathbf{B}) - trace((\mathbf{B}^T\mathbf{AR})^T + (\mathbf{B}^T\mathbf{AR})) + trace((\mathbf{AR})^T\mathbf{AR})\end{align*}
We also have that $trace(\mathbf{B}^T\mathbf{B}) = ||\mathbf{B}|| = 1$ (because the points have been normalised)
\newline
\\
and that $trace((\mathbf{AR})^T\mathbf{AR}) = ||\mathbf{AR}|| = ||\mathbf{A}|| = 1$ (because $\mathbf{R}$ is an isometry).
\newline
\\
Also, $trace((\mathbf{B}^T\mathbf{AR})^T) = trace(\mathbf{B}^T\mathbf{AR})$. Thus
\begin{align*}&trace(\mathbf{B}^T\mathbf{B}) - trace((\mathbf{B}^T\mathbf{AR})^T + (\mathbf{B}^T\mathbf{AR})) + trace((\mathbf{AR})^T\mathbf{AR}) \\ =&2 - 2 * trace(\mathbf{B}^T\mathbf{AR})\end{align*}
which means that we want to find $\mathbf{R}$ such that $trace(\mathbf{B}^T\mathbf{AR})$ is maximised. 
\newline
\\
Apply singular value decomposition to $\mathbf{B}^T\mathbf{A}$ to find $\mathbf{U}$, $\mathbf{V}$ and $\mathbf{W}$ such that
\[\mathbf{B}^T\mathbf{A} = \mathbf{V}\mathbf{W}\mathbf{U}^T\]
with $\mathbf{U}$ and $\mathbf{V}$ orthogonal matrices and $\mathbf{W}$ a diagonal matrix with non-negative entries.
\newline
\\
Then $trace(\mathbf{B}^T\mathbf{AR})$ = $trace(\mathbf{V}\mathbf{W}\mathbf{U}^T\mathbf{R})$ = $trace(\mathbf{X}\mathbf{W})$ where $\mathbf{X}$ is set as $\mathbf{X} = \mathbf{V}\mathbf{W}\mathbf{U}^T\mathbf{R}\mathbf{W}^{-1}$ and is orthogonal. Write:
\[\mathbf{X} = \left( \begin{array}{cc}
a & b\\
c & d
\end{array} \right)
\hspace{12pt}\mathbf{W} =  \left( \begin{array}{cc}
e & 0\\
0 & f
\end{array}\right)\]
Then $trace(\mathbf{XW}) = ae + df$. Since $\mathbf{X}$ is orthogonal, $a$ and $d$ have absolute value less than or equal to 1. This means that the maximum trace of $\mathbf{XW}$ is $e + f$ which is the trace of $\mathbf{W}$.
\newline
\\
Set $\mathbf{R}$ = $\mathbf{U}\mathbf{V}^T$. Then R maximises $trace(\mathbf{B}^T\mathbf{AR})$ because $trace(\mathbf{B}^T\mathbf{AR})$ = $trace(\mathbf{V}\mathbf{W}\mathbf{U}^T\mathbf{R})$ = $ trace(\mathbf{V}\mathbf{W}\mathbf{U}^T\mathbf{U}\mathbf{V}^T)$ = $trace(\mathbf{V}\mathbf{W}\mathbf{V}^T)$ = $trace(\mathbf{W})$
\newline
\\
Thus, we have shown that $\mathbf{R}$ = $\mathbf{U}\mathbf{V}^T$ minimises the Procrustes distance between $\mathbf{A}$ and $\mathbf{B}$ and the proof is complete. 
\end{proof}
Chapter 2 is now concluded. In the next chapter we report on the development of two novel measures of facial asymmetry. 
\newpage

\chapter{Developing two measures of facial asymmetry}\label{sec:chapterthree}
\section{About this chapter}
In this chapter we describe the design of two measures of asymmetry and the steps involved in their development. We start by presenting several software implementations of ASMs and AAMs and explaining how we decided which implementation to use as the basis of our measures of asymmetry. We test the selected implementation by training it for a test video and evaluating its fitting performance. We then go on to describe the development of the two measures in detail. The first measure relies only on shape data (i.e., coordinates of landmarks of the face) whereas the second measure uses both shape data and texture data (where texture data means pixel values of all points on the face). After describing the two measures, we compare their performance on the test video and decide which measure to prefer. We end the chapter by discussing the limitations of the preferred measure. 

\section{Software selection}\label{sec:softwareimp}
The first step was to select a software implementation of either Active Shape Models or Active Appearance Models that would be our starting point for developing measures of asymmetry. Several implementations were tried and tested, according to the following desiderata: 
\begin{itemize}
\item Ideally the software would implement AAMs to allow the possibility of using full appearance information (shape and texture) should appearance information prove to enable a better measure of asymmetry than shape information alone
\item The source code should be available to allow customisation of the software and complete understanding of its inner workings
\item The fitting should be accurate, reliable and capable of operating with enough speed to model videos of faces as well as single images of faces 
\end{itemize}
Seven different software implementations were considered and four of these were tested. Table~\ref{tab:software} summarises information about each of these implementations. Options 2, 5 and 6 were rejected because the source code could not be obtained. Option 1, which was in MATLAB, was capable of fitting shapes but was too slow to be used for videos. Options 3 and 4 worked well but were limited to ASMs and so were ultimately rejected. 
\begin{table}[htcp]\footnotesize
\centering
\begin{tabular}{|p{0.3cm}|p{1.5cm}|p{1.5cm}|p{1cm}|p{1.2cm}|p{5.1cm}|}
\hline
& Author & Language & ASM or AAM? & Source Available? & Verdict \\ \hline
1 & Ghassan Hamarneh & MATLAB & ASM & Yes & Too limited. No facility for tracking videos, without extra coding. Very slow tracking.\\ \hline
2 & Tim Cootes & C++ & AAM & No & Source code was not available so this option was ruled out.\\ \hline
3 & Stephen Milborrow & C++ & ASM & Yes & Worked well but ultimately rejected in favour of option 7 which was more versatile.\\ \hline
4 & Yao Wei & C++ & ASM & Yes & By the same author as option 7, but limited to ASMs whereas 7 implements AAMs as well.\\ \hline
5 & Mikkel B. Stegmann & C++ & AAM & No & Although C++ source was meant to be available for download, it could not be obtained.\\ \hline
6 & George Papandreou & MATLAB and C++ & AAM & No & Author did not respond to email requesting source code.\\ \hline
7 & Yao Wei & C++ & AAM & Yes & Selected this. Fitted images and videos. Performed both training and fitting.\\ \hline
\end{tabular}
\caption{Candidate software implementations of ASMs and AAMs (the associated URLs can be found in Appendix~\ref{sec:appendixone}).}\label{tab:software}
\end{table}

% put the links in the appendix
%http://www.cs.sfu.ca/~hamarneh/software/asm/index.html
%http://www.mathworks.com/matlabcentral/fileexchange/26706-active-shape-model-asm-and-active-appearance-model-aam
%http://personalpages.manchester.ac.uk/staff/timothy.f.cootes/software/am_tools_doc/index.html
%http://www.milbo.users.sonic.net/stasm/
%http://code.google.com/p/asmlibrary/
%http://www2.imm.dtu.dk/~aam/
%http://cvsp.cs.ntua.gr/software/AAMtools/
%http://code.google.com/p/aam-library/

The AAM library by Yao Wei (number 7 in Table~\ref{tab:software}) was selected because it met all of the desiderata: the source code was available; it allowed access to both appearance information and shape information; it allowed both training \emph{and} fitting of models; it was much faster than the MATLAB implementation that was tested (option 1); and it was recommended by two PhD students at the University of Bristol who had used it in their research (Alexander Davies and Xiao Wang). 

The library was compiled using the Eclipse Integrated Development Environment in Ubuntu 11.04. It was necessary to compile OpenCV (Open Source Computer Vision Library) beforehand. Version 2.2 of OpenCV was used.

\section{Testing the AAM library} 

The first test of the AAM library was to train it on a face in a test video, and see if it would track the face reliably. The video used was provided by the Psychology Department at the University of Bristol. A three-minute segment was extracted from the video using a tool called ffmpeg. The segment was chosen to include a variety of facial expressions; the subject is seen talking, smiling and laughing and there is some head movement and rotation. To increase speed of fitting, the resolution was reduced using ffmpeg to 640 by 360 pixels from the original resolution of 1920 by 1080 pixels.

\subsection{Training}\label{sec:training}

To train the AAM it was necessary to select several frames from the video that represented a range of expression and would allow the model to fit all frames in the three-minute segment. This could have been done by manually selecting frames on the basis of visual inspection, but an alternative methodology was adopted. To select the first cohort of frames a small C++ program was written that selected the frames using the principle of a polyline algorithm. A polyline algorithm is essentially a method for simplifying a polyline. A three-minute segment of video - at 25 frames per second - consists of 4,500 frames. Each frame is an image of 640 by 360 pixels, and each pixel has three integer values; one for the red value, one for the green value and one for the blue value. Each frame can thus be represented as a 691,200 ( = 640 * 360 * 3) dimensional vector, with each value in the vector between 0 and 255. A three-minute video can therefore be thought of as a set of 4,500 points in 691,200-dimensional space. If these points are connected, then the result is a polyline. The polyline algorithm simplifies the polyline by selecting a number of vertices that represent the ``significant features'' of the polyline.

The implementation of the polyline algorithm is straightforward. Choose an integer value called the ``Tolerance''. The higher the tolerance, the fewer points that will be selected by the algorithm (i.e., the fewer frames that will be selected from the video). The tolerance value has to be chosen by trial and error. Pick an arbitrary value and run the program and see how many points it returns. If too few, then decrease the tolerance and run again; if too many, then increase the tolerance and run again. The program loads the first image from the video and sets it as the base image. It then steps through the video frame by frame, each time calculating the Euclidean distance between the base image and the given frame (recall that each frame is a 691,200 dimensional vector, so we can calculate Euclidean distances between frames). If the distance is greater than the tolerance then it saves a copy of the frame, and sets the frame as the base image. If the distance is less than the tolerance then the frame is not saved and the base image does not change.

\begin{figure}[!h]
\centering
\includegraphics[scale=0.28]{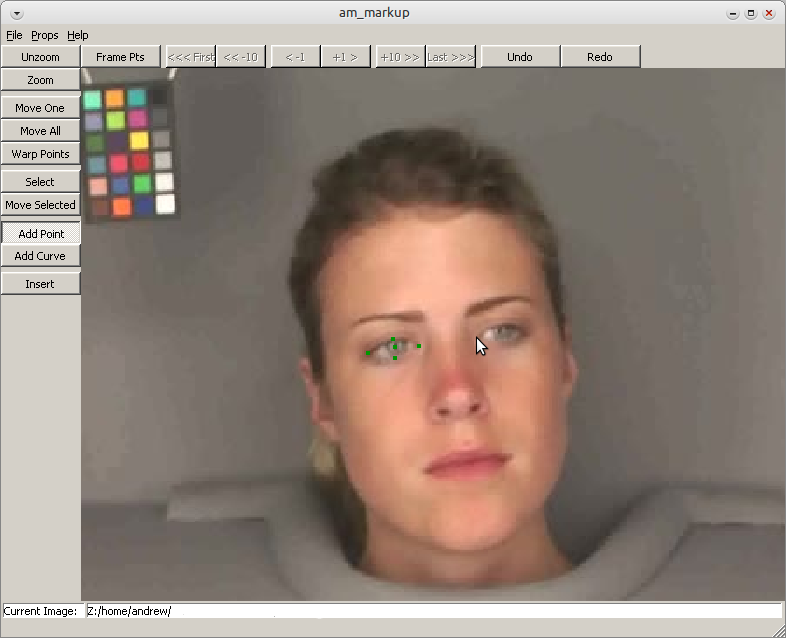}
\caption{Annotating an image using the markup tool written by Professor Tim Cootes and available on his website.}\label{fig:markup}
\end{figure}

\begin{figure}[!h]
\centering
\includegraphics[scale=0.45]{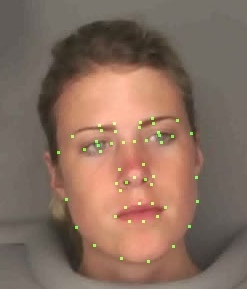}
\caption{Example of a frame with 50 landmarks marked.}\label{fig:girl_annotated}
\end{figure}

Using this method, five frames were selected from the video. These five were annotated, each with 50 landmark points, marking the significant features of the face. Annotation was done using a free tool built by Professor Tim Cootes available on his website (the URL can be found in Appendix~\ref{sec:appendixone}). The tool takes an image file, and allows the user to mark landmarks by hand (Figure~\ref{fig:markup} shows the tool in action). The landmarks are then saved in a .pts file, which is a text file that stores the x and y coordinates for each landmark. Figure~\ref{fig:girl_annotated} shows one of the annotated frames.

After annotating five images, the model could be built using the AAM library. To do this, five images and five corresponding .pts files (containing the coordinates of the landmarks) are needed. These are placed in a folder that the AAM library can access, and after the model builder is run, a model file is created. This model file is used in the fitting stage. 

\subsection{Fitting}\label{sec:aamlibfit}

\begin{figure}[htp]
\centering
\includegraphics[scale=0.5]{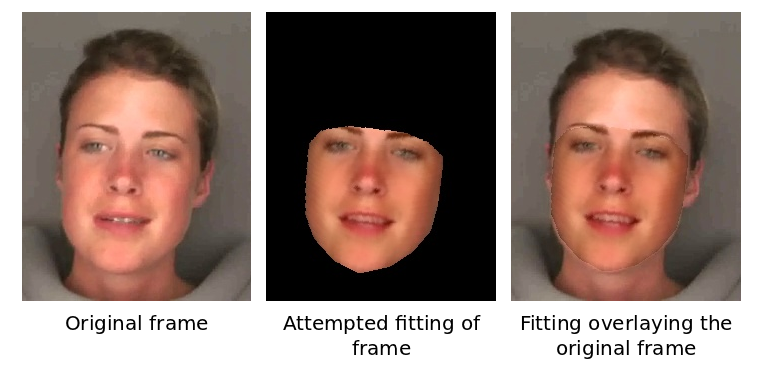}
\caption{Example of fitting a frame with a model built from 5 training images.}\label{fig:5iter_girl}
\end{figure}

The AAM library's fitting program takes a video and model file as input, and steps through the video frame by frame attempting to fit the face in each frame. It outputs a video of the synthesised face built by the model. In Figure~\ref{fig:5iter_girl} an example frame is shown.
The left-most image shows the original frame, and the middle image shows the attempt to fit the frame using a model created with 5 training images (each with 50 landmarks annotated). The right-most image shows the fitting overlaying the original frame. To measure the performance of the fitting, an OpenCV function calculating the L1 norm between the original frame and the fitted frame (i.e., left-most and right-most images in Figure~\ref{fig:5iter_girl}) was used. As we have already seen, images can be expressed as vectors. If im1 and im2 are two colour images (each of resolution 640 by 360 pixels) then we can write them both as 691,200-dimensional vectors:
\[\mathbf{im1} = (r_{1}, g_{1}, b_{1}, ... ,r_{230400}, g_{230400}, b_{230400})\]
and
\[\mathbf{im2} = (r'_{1}, g'_{1}, b'_{1}, ... ,r'_{230400}, g'_{230400}, b'_{230400})\]
The L1 norm for $\mathbf{im1}$ and $\mathbf{im1}$ is then defined as:
\[\sum\limits_{i=1}^{230400} |r_{i} - r'_{i}| + |g_{i} - g'_{i}| + |b_{i} - b'_{i}|\]
The L1 norm, then, is simply the sum of absolute differences between corresponding pixel values across the two images. The lower the L1 norm between the original frame and the fitted frame, the better the fitting performance. 

Calculating the L1 norm for each frame meant that the performance for one frame could be compared to the performance for another, and also that the performance of one model could be compared to the performance of another model, by looking at each model's performance averaged across all frames. The frame displayed in Figure~\ref{fig:5iter_girl} represents a fitting with average performance (i.e., the median performance for all frames in that video). It can be seen that the fitting is not particularly good. The colours do not look right and the shape of the mouth is wrong. In addition, the eyes point in a different direction to their direction in the original frame. However, since this model was the result of only 5 training images, there is scope to improve the model, as we shall see. 

Two modifications were made to the AAM library's fitting code. One was to add a function that produced the frame with the fitting overlaying the original frame (for example, the right-most image in Figure~\ref{fig:5iter_girl}). This was achieved using OpenCV functions. The second modification was to add a colour correction function, which significantly improved the colour of the fitting. The colour correction function starts by considering the red colour channel. It calculates the average red pixel value across the original frame and subtracts the average red pixel value across the fitted frame. The result is then added to each red pixel value of the fitted frame (if this results in a red pixel value of less than 0, then 0 is chosen; if it results in a red pixel value of greater than 255, then the value is set to 255). The same is then done for the green and blue colour channels. The function made no noticeable difference to speed of fitting and substantially improved fitting performance.  

\begin{figure}[htp]
\centering
\includegraphics[scale=0.55]{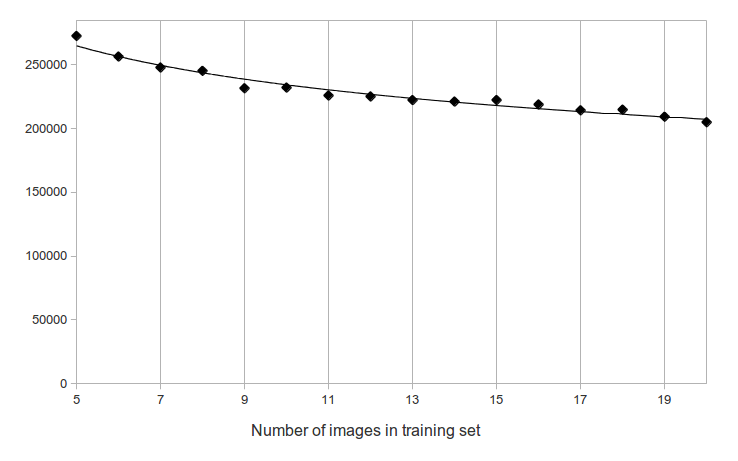}
\caption{The chart shows average fitting error across frames on the y-axis against number of training images used to build the model on the x-axis. A trend line has been added.}\label{fig:chart_fit}
\end{figure}

We expected that increasing the number of training images would improve fitting performance, but according to diminishing returns (\cite{gmb05} provided support for this expectation). Having created a model from 5 training images, and having attempted to fit this model to the test video, it was possible to see which frame fitted least well by looking at the L1 norm between each original frame and its corresponding fitted frame. The frame which was fitted least well was annotated and added to the original 5 frames to build a model with 6 images. This model was then fitted to the video; its average performance was recorded (i.e., average L1 norm across all frames); and the worst performing frame was used to create a model with 7 training images. This process was repeated until a model had been built with 20 training images. This allowed us to plot a chart indicating the impact on fitting performance of the number of training images, for the test video. Figure~\ref{fig:chart_fit} shows average fitting error (i.e., the average L1 norm between each original frame and its corresponding fitted frame for the video) against the number of training images used. In all cases, 50 landmarks were used in the training set. It can be seen that earlier additions to the number of training images make more substantial improvements to performance than later additions; for example, increasing the number of training images from 5 to 6 has more of an impact on performance than increasing the number from 19 to 20. A trend line has been plotted for the points. We can see from the trend line that 20 training images is by no means the optimum number of images for this test video, but for the present purpose of establishing that AAM library can achieve a good fitting, we shall see that it is enough. 

\begin{figure}[!h]
\centering
\includegraphics[scale=0.9]{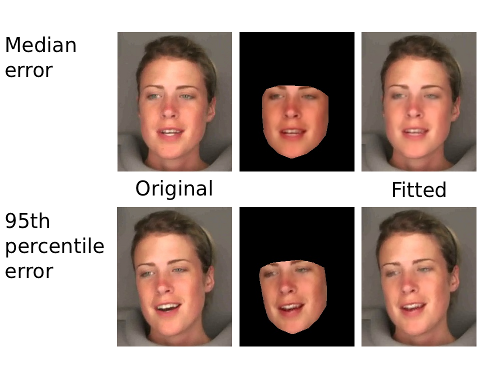}
\caption{The fitting performance - for two frames - of the model created using 20 training images.}\label{fig:compare}
\end{figure}

Figure~\ref{fig:compare} shows the performance of the model created using 20 training images for two frames. The first frame is the frame whose fitting error is the median value of the set of all fitting errors for the frames in the video. It can therefore be viewed as representative of the average performance of the model for the test video. The second frame has a fitting error in the 95th percentile. In other words, 95\% of frames have fitting error less than or equal to the error for the second frame.  

Inspection by eye of the fitted frames in Figure~\ref{fig:compare} shows that the result achieved by using 20 training images is a good one. On the top row (median error), all facial features are faithfully represented. The only noticeable difference between the original face and the synthesised face is a lack of sharpness in the latter. In addition, more lower teeth are exposed in the synthesised face than the original, but this is only a subtle difference. Since 50\% of frames were fitted better than this frame, it was clear that the AAM was performing well. Also notice that the colours of the synthesised model have been greatly improved in comparison to Figure~\ref{fig:5iter_girl}. This was a result of the colour correction function. 
On the bottom row we can see a frame that is fitted less well, but the result is still satisfactory. Again there is a lack of sharpness in the synthesised image and the shape of the mouth appears slightly different. In addition, the eyes in the synthesised image appear to be directed slightly differently. However, all in all, performance in the bottom row is also good, and 95\% of frames in the test video showed better performance.

By this stage it was deemed that the AAM library was sufficient for the purpose of building a measure of asymmetry, because it was able to accurately model faces in videos and locate facial features. The next step was to build on the library a means of measuring facial asymmetry. 

\section{Two measures of facial asymmetry}

It turned out that two measures of asymmetry were developed. The first measure was simpler because it only used shape information to measure asymmetry. The second measure built on the first measure by using texture information in addition to shape information. In this section we describe the two measures, compare the two and explain why the first was ultimately preferred to the second. 

\subsection{Measure One: Using shape information only}\label{sec:measureone}

To develop the first measure of facial asymmetry, we started by creating a short test video with the following characteristics:  

\begin{itemize}
\item First, the face was pointed directly to the camera, with minimal left or right rotation. The reason for this was that left or right rotation would distort a 2-dimensional measure of asymmetry (we show why this is in Section~\ref{sec:posevariation}).
\item Second, the video showed the face with neutral expression punctuated by asymmetric expressions such as raising one eyebrow or raising one side of the mouth. 
\end{itemize}

\begin{figure}[htp]
\centering
\includegraphics[scale=0.58]{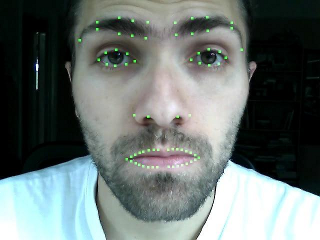}
\caption{An annotated training image from the test video.}\label{fig:test_annotated}
\end{figure}

The AAM was trained in the usual way and training images were selected as above. That is, first a polyline algorithm was used to select 5 images, and then further images were added to improve tracking where necessary. Once a satisfactory fitting of the test video was achieved, we could begin to develop the measure. In the training phase we had used 68 landmarks on each image (see Figure~\ref{fig:test_annotated} for an example of an annotated training image).
The AAM library stored shape data during fitting in a matrix. This meant that we could access this matrix for each fitted frame and extract the x and y coordinates for the 68 points. If the fitting was accurate then these coordinates would represent the same landmarks of the face that were annotated in the training phase. We could then use the coordinates of these points to measure the degree of asymmetry of the face in the given frame using Procrustes analysis. In the following we break down the steps required to perform this calculation on a given frame.

\subsubsection{Step One}

Given a frame, extract the shape data for the fitting and separate the 68 points into those that belong to the left side of the face and those that belong to the right side. To separate the points we calculate the median value of the x-coordinates of the 68 points. Points with x-coordinate less than the median are deemed to be for the left side of the face; points with x-coordinate greater than the median are deemed to for the right side of the face. Separating the points in this way results in two subsets of points, each with 34 points. (Using the median of the x-coordinates for the separation relies on the face being relatively upright in each frame; this assumption was reasonable because of the way we opted to film the videos. See the later section on pose variation (\ref{sec:posevariation}) for more detail.)

\begin{figure}[htp]
\centering
\includegraphics[scale=0.52]{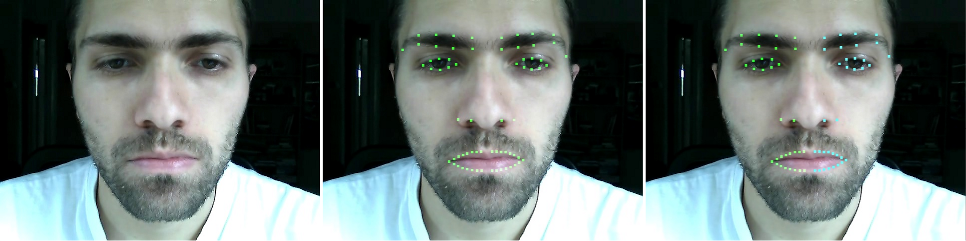}
\caption{Plotting shape data for a fitted frame and separating according to side of face.}\label{fig:separated}
\end{figure}

Figure~\ref{fig:separated} shows three images for a frame of the test video. The left-most is the frame itself. The middle shows the fitted points overlaying the frame, and the right-most frame shows the same points separated into left-side points (coloured green) and right-side points (coloured blue).

\subsubsection{Step Two}

Once we have the two sets of points representing each side of the face, and they are in suitably dimensioned matrices (i.e., 2 columns by 34 rows, for the (x,y) coordinates of each of the 34 points) we can perform Procrustes analysis to determine the rotation, reflection and translation that brings the right-side set of points closest to the left-side set, where the notion of distance is Procrustes distance. Procrustes analysis is applied to the two matrices containing the coordinates of the points. There are a number of steps, which we now detail. Functions were written to perform these steps in C, using OpenCV matrix manipulation functions where appropriate.

\begin{itemize}
\item First, both matrices are centred on the origin by calculating their centroids and translating by the negative of their respective centroids (the centroid of a set of n points, ${(x_{i}, y_{i})}$, means the point ($\overline{x}, \overline{y})$ where $\overline{x} = \frac{1}{n}\sum\limits_{i=1}^{n} x_i$ and $\overline{y} = \frac{1}{n}\sum\limits_{i=1}^{n} y_i$ are the means of the set of x-coordinates and y-coordinates respectively).
\[(x_{i}, y_{i}) \rightarrow (x_{i} - \overline{x}, y_{i} - \overline{y})\]
\item Second, both matrices are normalised so that the sum of the squares of their values is 1.
\[(x_{i}, y_{i}) \rightarrow (\frac{x_{i}}{C}, \frac{y_{i}}{C})\] where $C = (\sum\limits_{i=1}^{n} {x_i}^2 + {y_i}^2)^{\frac{1}{2}}$ 
\item Third, the first matrix (that represents the left-side set of points) is transposed and multiplied by the second matrix (that represents the right-side set of points). Singular value decomposition is applied to the resulting 2 by 2 matrix, to decompose the matrix into a product of three 2 by 2 matrices; call these matrices $\mathbf{L}$, $\mathbf{D}$ and $\mathbf{M}^T$ where $\mathbf{L}$ and $\mathbf{M}$ are orthogonal matrices and $\mathbf{D}$ is the diagonal matrix. 
\item Fourth, the orthogonal matrix that represents the rotation and reflection that minimises Procrustes distance (call it $\mathbf{T}$) is found by taking the product of $\mathbf{M}$ with the transpose of $\mathbf{L}$. I.e., $\mathbf{T} = \mathbf{M}\mathbf{L}^T$. We have proved that $\mathbf{T}$ is the minimising matrix in Section~\ref{sec:procrustes}.
\item Fifth, the translation component of the transformation that minimises the Procrustes distance is found by applying $\mathbf{T}$ to the original centroid of the right-side set of points and subtracting the result from the original centroid of the left-side set of points (the centroids were calculated in the first step). 
\end{itemize}

\subsubsection{Step Three}

\begin{figure}[!h]
\centering
\includegraphics[scale=0.45]{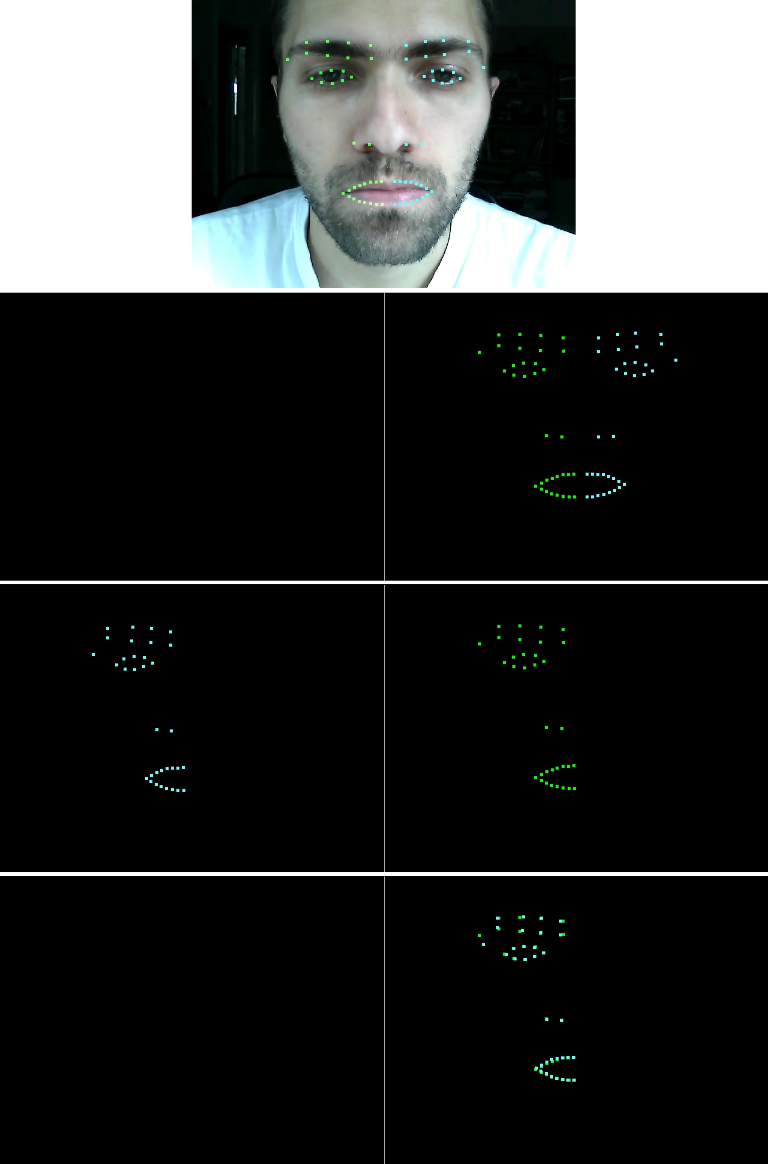}
\caption{The top row shows the frame with left-side and right-side points fitted. The second row simply shows these points on a black background (the vertical white line in the centre of this row represents the y-axis (i.e., the line x = 0)). The third row shows the left-side points and the right-side points after the latter have had the rotation and reflection applied. The fourth row shows the left-side points and the right-side points after the rotation, reflection and translation have been applied. The Procrustes distance between corresponding points in the fourth row is the measure of asymmetry for the frame. It was found to be 20.52.}\label{fig:measureone}
\end{figure}

Given a frame, and the reflection, rotation and translation that minimises the Procrustes distance, we apply the transformation to the right-side set of points and measure the Procrustes distance between the resulting set of points and the left-side set of points. The distance is our measure of asymmetry for the frame. The larger the distance, the larger the asymmetry. In this way we were able to develop a program that - when given a video of a face and a model trained for that face - could put a frame-by-frame measure of facial asymmetry on the video. In Figure~\ref{fig:measureone} we show this measure at work on a neutral and relatively symmetric face. We can see that after rotating, reflecting and translating the right-side set of points they are fairly close to the left-side set of points. The Procrustes distance for this frame was found to be 20.52 (to two decimal places).

\begin{figure}[!h]
\centering
\includegraphics[scale=0.34]{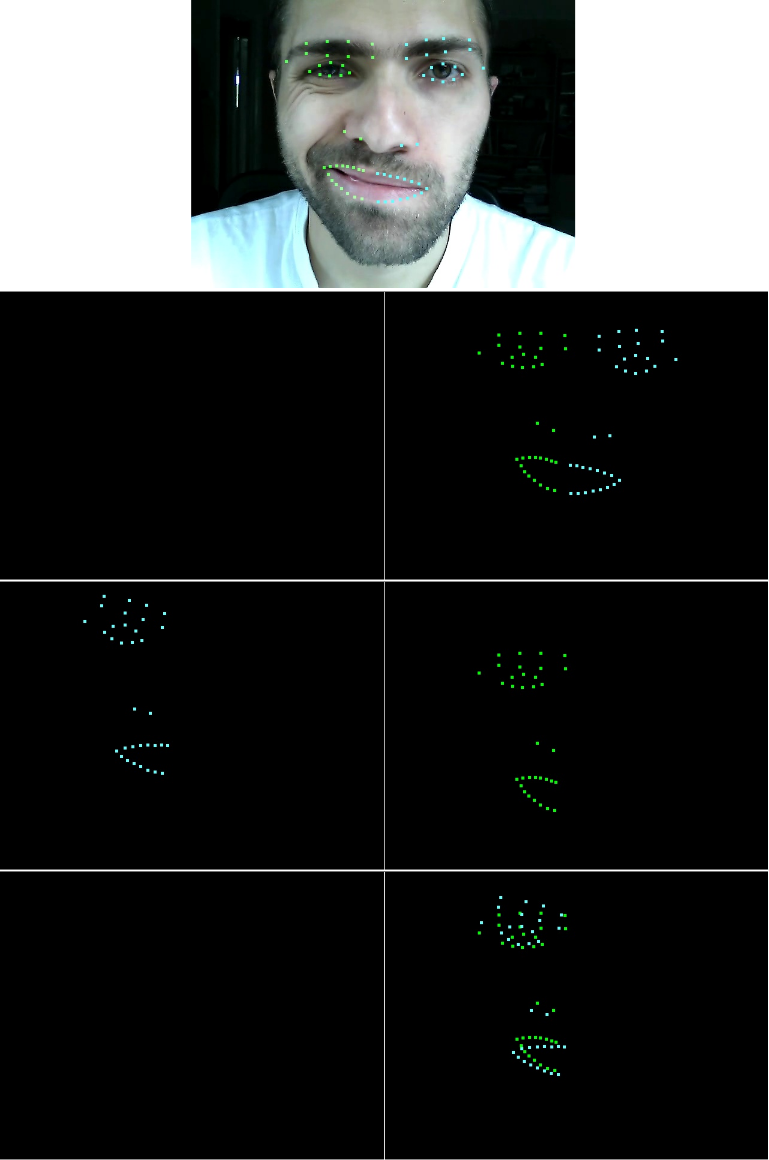}
\caption{These images show the first measure at work on a deliberately asymmetric face. The methodology is the same as for Figure~\ref{fig:measureone}. The Procrustes distance for this frame was found to be 94.30, indicating a high level of asymmetry.}\label{fig:measureone2}
\end{figure}

In Figure~\ref{fig:measureone2} we show the measure at work on a deliberately asymmetric face. The Procrustes distance for this frame was found to be 94.30 (to two decimal places).

\subsection{Measure Two: Using shape \emph{and} texture information}\label{sec:measuretwo}

The second measure of asymmetry developed used both shape \emph{and} texture information. The idea here was, given a frame, fit the frame using the AAM (so we would have both shape and texture information); determine an axis of symmetry for the fitting; reflect the texture information from one side of the face across to the other side of the face; measure the average absolute difference in corresponding pixel values between the reflected side and the unreflected side.

\begin{figure}[htp]
\centering
\includegraphics[scale=0.9]{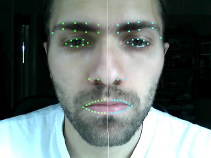}
\caption{A fitted frame from the test video with left-side and right-side points marked, and the axis of symmetry drawn in white.}\label{fig:axissymmetry}
\end{figure}

The first step was to decide how to determine an axis of symmetry for each frame. It was decided to take the following approach. First, extract the shape data and centre the points on the origin by calculating their centroid and subtracting the centroid from each pair of coordinates. Second, split the points into those for the left-side and those for the right-side (as with the first measure). Third, apply Procrustes analysis to determine the transformation that, when applied to the right-side points, minimised the Procrustes distance to the left-side points. As already seen, Procrustes analysis yields an orthogonal matrix representing the rotation and reflection. We can use this matrix to infer the axis of symmetry. The matrix has the form: \[ \left( \begin{array}{ccc}
-cos \theta & sin \theta \\
sin \theta & cos \theta\end{array} \right)\] 
This is equivalent to reflecting in the straight line y = mx where \[m = \frac{cos\frac{\theta}{2}}{sin\frac{\theta}{2}}\]
Figure~\ref{fig:axissymmetry} shows a frame from the test video with the shape data overlaying the image and with the axis of symmetry (calculated in the way described) drawn as a white line. 

\begin{figure}[htp]
\centering
\includegraphics[scale=0.35]{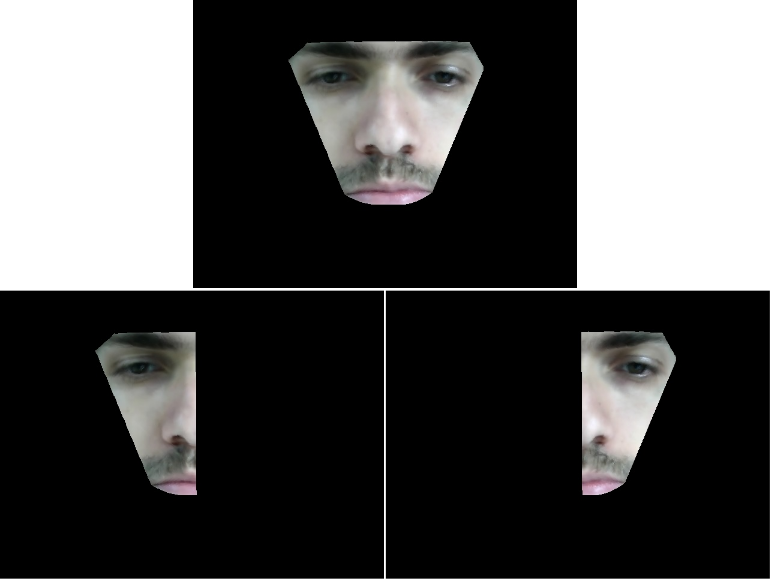}
\caption{The top row shows the fitting provided by the AAM model. The second row shows it split into left and right sections according to the calculated axis of symmetry.}\label{fig:texturesplit}
\end{figure}

The second step was to split the texture information according to the axis of symmetry. This provides a fit to the left side and a separate fit to the right side. Figure~\ref{fig:texturesplit} indicates what this split looks like. 
The third step is to reflect one side of the fitting across the axis of symmetry. Without loss of generality, the right side was reflected. To perform the reflection, we used the fact that reflection of a point (x', y') in a line y = mx takes the point to: \[(\frac{2my' - (m^2 - 1)x'}{1 + m^2}, \frac{(m^2 - 1)y' + 2mx'}{1 + m^2})\](See Appendix~\ref{sec:reflection} for the proof.)

\begin{figure}[!h]
\centering
\includegraphics[scale=0.23]{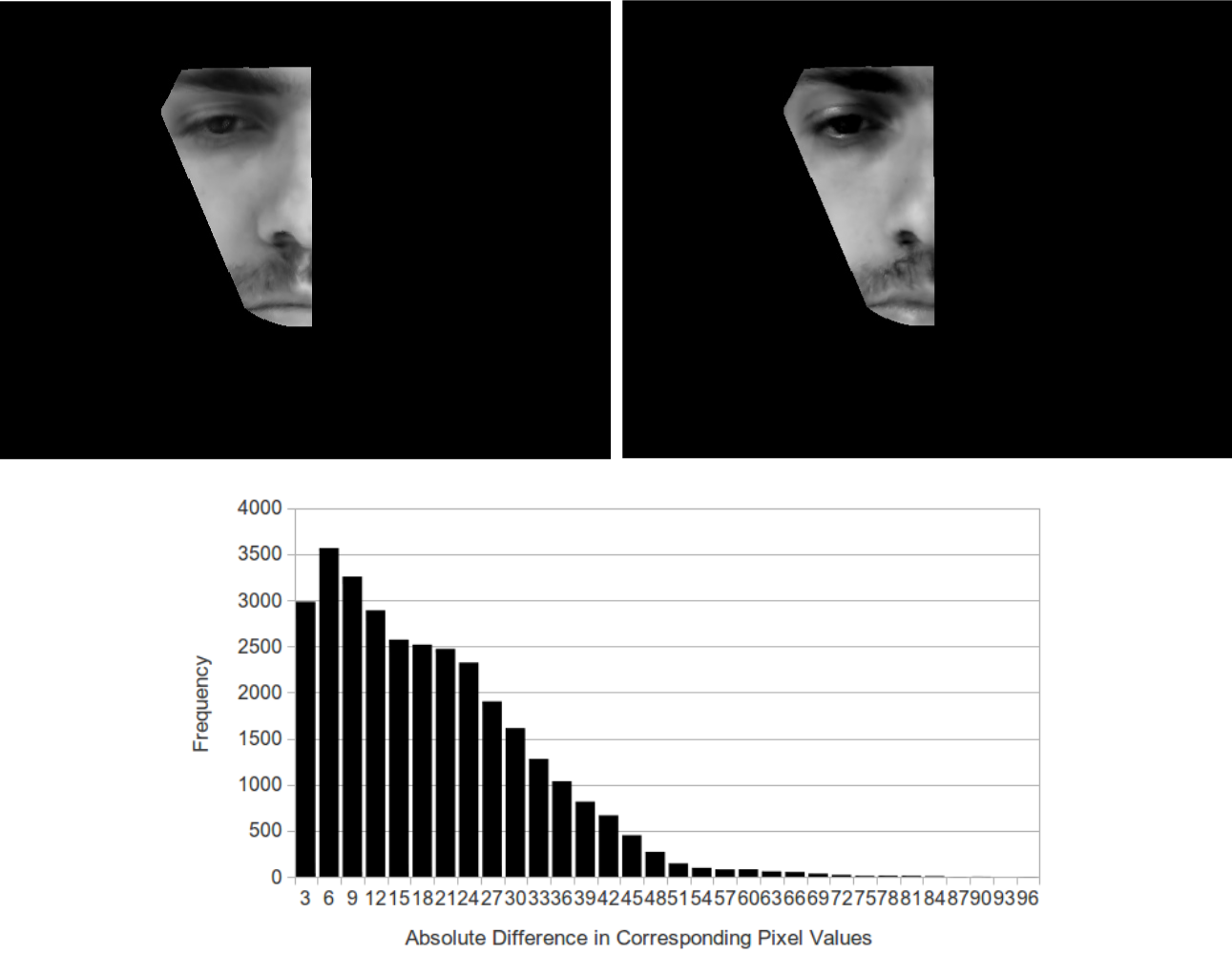}
\caption{The left image shows the left side of the face, unadjusted. The right image shows the right side of the face, reflected in the axis of symmetry. Below the images is a histogram showing the distribution of absolute differences between corresponding pixel values for the two images. The mean absolute difference is 18.38 and the median is 16. Our measure of asymmetry hypothesises that the higher these numbers, the greater the facial asymmetry.}\label{fig:grayscale}
\end{figure}

\begin{figure}[!h]
\centering
\includegraphics[scale=0.40]{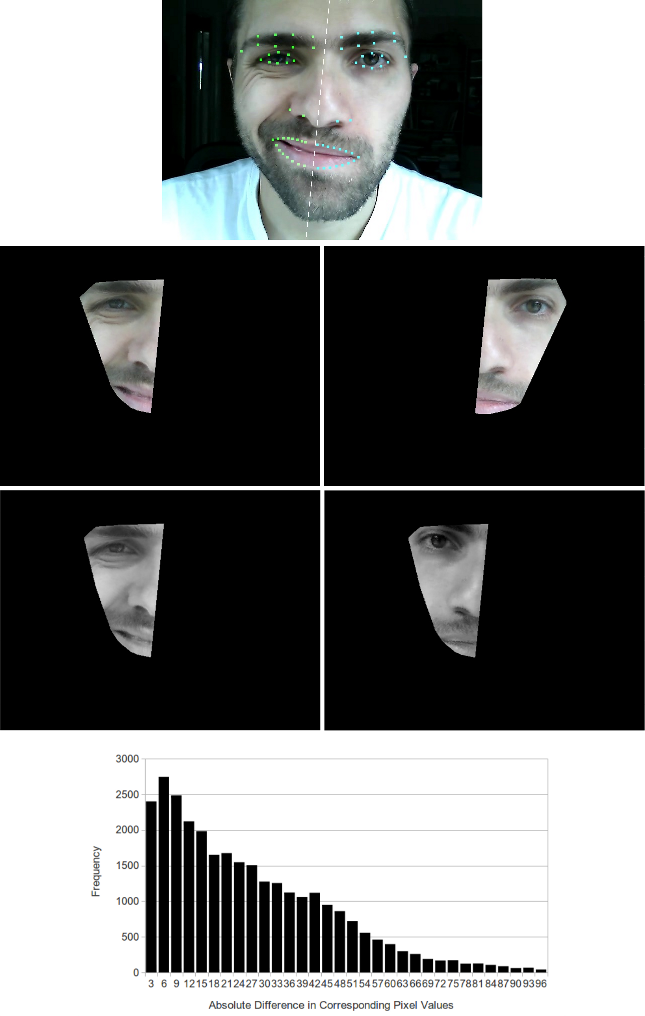}
\caption{The top row shows a deliberately asymmetric face with the axis of symmetry drawn. On the second row we see the left and right fittings of the face. On the third row we see the left side of the face, and the right side reflected across the axis, both in grayscale. The histogram shows the distribution of absolute differences between corresponding pixel values for the two grayscale images. The mean absolute difference is 24.82 and the median is 20. Comparing this histogram to the histogram in Figure~\ref{fig:grayscale} reveals a longer tail, indicative of greater difference between the grayscale images.}\label{fig:histo2}
\end{figure}

The final step is to convert both the left side and the reflected right side to grayscale and to normalise them with respect to average pixel value. The reason we do this is to reduce the likelihood of error in our measure of asymmetry caused by one side of the face being more illuminated than the other due to uneven lighting during recording. (To perform the normalisation we increment (or decrement) the pixel values of the reflected right side by a constant so that its average pixel value is the same as the average pixel value of the left side.) Finally, we crop each image so that their shapes are identical, keeping as much of each image as possible. The result of these operations is shown in Figure~\ref{fig:grayscale}. The mean absolute difference between corresponding pixel values for the two grayscale images is 18.38 and the median is 16.
In Figure~\ref{fig:histo2} we apply the same measure to a deliberately asymmetric face. The mean absolute difference in pixel value in this image is 24.82 and the median is 20. This agrees with our expectation that the higher the asymmetry the higher the average absolute difference between corresponding pixel values for the unreflected left side and the reflected right side.  

The second measure can be subdivided into two distinct but closely related measures; we can either use the \emph{mean} absolute difference in pixel values or the \emph{median}. The reason why we may want to use the median rather than the mean is that the mean - as a measure of the average of a distribution - can be skewed if the distribution is skewed. If, for example, there are a minority of pairs of pixels where the absolute difference is very high, the mean absolute difference may be high even though the two grayscale images are similar. To see if using the median produced a significantly different measure to using the mean, we used both methods to record the degree of asymmetry for each of the 1,250 frames of the test video. We then calculated the Pearson correlation coefficient between the mean absolute pixel difference and the median absolute pixel difference. (The Pearson correlation coefficient between two variables is their covariance divided by the product of their standard deviations - the maximum possible value is 1 and the minimum possible value is -1). The Pearson correlation turned out to be 0.93 indicating a strong degree of dependence between the variables. We also plotted both variables against frame number and found that there was no interesting difference between using the median and using the mean. 

\subsection{Comparing the measures}\label{sec:comparison}

\begin{figure}[!h]
\centering
\includegraphics[scale=0.53]{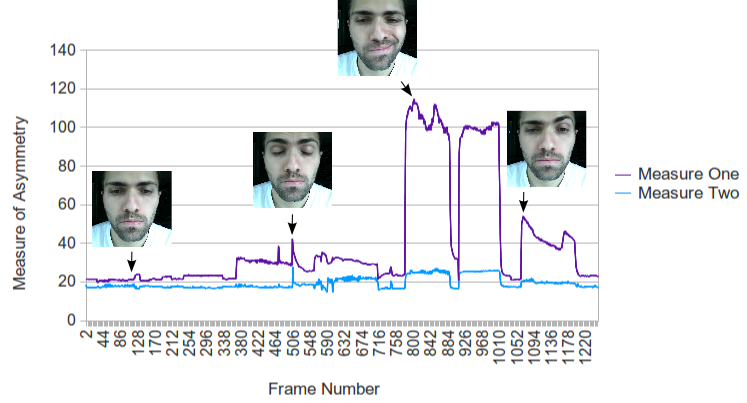}
\caption{Chart showing measured facial asymmetry on the y-axis against frame number on the x-axis for the test video. The magenta line is for Measure One and the blue line is for Measure Two.}\label{fig:comparemeasures}
\end{figure}

We now have two different measures of facial asymmetry. The first was described in Section~\ref{sec:measureone} and relies only on shape information; call this ``Measure One''. The second was described in Section~\ref{sec:measuretwo} and uses both shape and texture information to find the mean absolute difference between pixel values for one side of the face and the other; call this ``Measure Two''. To compare these two measures we performed both on all 1,250 frames of the test video and plotted the results in Figure~\ref{fig:comparemeasures}. The magenta line shows asymmetry by frame number according to Measure One and the blue line shows asymmetry by frame number according to Measure Two. The Pearson correlation between the two measures was calculated to be 0.90 indicating a high degree of dependence between the variables. 

On the graph we have displayed four frames at points of interest. The first frame (left-most) is a neutral expression. Both measures record a relatively low level of asymmetry. The second frame is a neutral expression with the eyebrows raised and the eyes closed. Both measures record a small jump in asymmetry at this point, which, as we shall see again in Section~\ref{sec:analysis}, is due to the left eyebrow arching more than the right and the left eyelid lowering more than the right during blinking. The third frame is a deliberately asymmetric frame where the left corner of the mouth is deliberately raised. Both measures record a jump in asymmetry, but the jump is much greater for Measure One. Measure Two records a level of asymmetry no greater than the level that it records for the second frame, which seems wrong. The fourth frame is another deliberately asymmetric frame where the left eyebrow is raised. Again, both measures record an increase in asymmetry, but the jump is greater for the first measure. 

On the basis of this video and other considerations we decided to prefer Measure One and discard Measure Two. The reasons for this were that:  
\begin{itemize}
\item Although the two measures were highly correlated for the test video (with a Pearson correlation coefficient of 0.90) Measure One performed better and more closely matched our expectations than Measure Two. In particular, Measure One correctly identifies that frames 3 and 4 display greater asymmetry than frame 2, whereas Measure Two records frame 2 as the most asymmetric point of the video.
\item Measure One is customisable because it allows the user to choose which parts of the face should figure in the measure of asymmetry. To include a part of the face in the measure, the user simply needs to annotate it during the training phase. Furthermore, if the user wants certain parts of the face to weigh more in the measure of asymmetry, then the user simply has to add extra training points to those parts of the face. Measure Two is not customisable which means that the user cannot choose to exclude any asymmetry of pixel values from the measure. For example, if the subject looks to one side this will create asymmetry that Measure Two cannot ignore. Measure One can either include or ignore the asymmetry depending on whether the user annotates the pupils of the eyes in the training stage. 
\item Measure Two can be affected by lighting variation during recording of the video whereas Measure One cannot. 
\end{itemize}  
For these reasons, we discard Measure Two at this point. In the next section we look at the limitations of Measure One.\section{Limitations of the measure}

\subsection{Fitting noise}\label{sec:fittingnoise}

There are two main limitations of our measure (i.e., Measure One). The first is that it is only as good as the shape data provided by the fitting. If the shape data do not correctly locate the points of the face that they are meant to locate, then the measure of asymmetry will be incorrect. There are two methods that we can use to check that the fitting reaches a certain standard.  However, since both of these methods can only show that the fitting is accurate up to a certain point, our measure of asymmetry necessarily has room for a small degree of error.

The first way to verify that the fitting reaches a certain standard is to watch the video with the shape data overlaying the frames. If points are incorrectly located then the viewer will typically spot this. Experimentation showed that the video can be watched back at double speed and the viewer can still easily identify when points are incorrectly located (and can respond by adding extra images to the training set). However, checking the shape data by eye necessarily limits the degree of accuracy that we can be sure of achieving (and is time-consuming). The viewer is unlikely to spot that a point is out of place by just two or three pixels. And a viewer is more likely to miss a misplaced point if they need to check a lot of frames.

Recall that in Section~\ref{sec:aamlibfit} we showed that we can measure the fitting error for a given frame by taking the L1 norm of the difference between the original frame and the fitted frame. We would like an analogous way of automatically measuring the fitting error of the shape data alone. However, since we have no way of accessing the shape data, independently of using the fitting, we must rely on checking the fitting by eye.

The second way that we can verify that the fitting reaches a certain standard is as follows. Remember that the first measure involves rotating, reflecting and translating the right-side set of points to minimise the Procrustes distance to the left-side set of points. If there are 34 points in each side's set of points then we can record the 34 distances between corresponding points for each frame. We can then sort these distance (for all frames) and look at the frames that have the largest distances. The reason is that the largest distances \emph{could} be due to fitting errors and so they should be checked. If the largest distances are due to facial asymmetry rather than fitting error, then we can be confident that there are no other substantial fitting errors in the video. This process was completed for the test video, and we found that the largest distances were due to asymmetry and not fitting error.

In summary, our measure of asymmetry is only as good as the accuracy of the fitted shape data, and we can only show that the fitting is accurate up to a certain point, so our measure of asymmetry necessarily has room for a small degree of error.

\subsection{Pose variation}\label{sec:posevariation}

\begin{figure}[htp]
\centering
\includegraphics[scale=0.76]{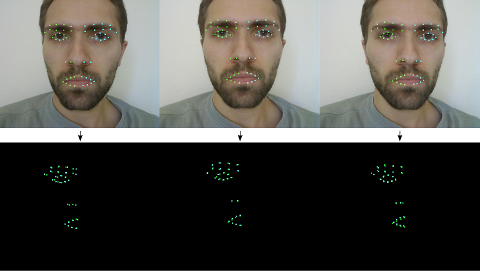}
\caption{Three images illustrating that our measure of asymmetry is affected by lateral head rotation. Below the images we can see the landmarks from the left-side of the face and the landmarks from the right-side of the face after reflection, rotation and translation. Recall that Procrustes distance between corresponding points is our measure of asymmetry. The Procrustes distance for the left image is 31.57; the Procrustes distance for the middle image is 19.19; and the Procrustes distance for the right image is 24.20.}\label{fig:posevariation}
\end{figure}

The second limitation of our measure is that rotational, lateral movement of the head will affect the measure of facial asymmetry even though the facial expression is constant and so the actual level of asymmetry is unchanged. Figure~\ref{fig:posevariation} illustrates this point. The left-most image shows the head rotated to the right by a small angle and the right-most image shows it rotated to the left by a small angle. The centre image has minimum rotation and is intended to be straight on. The measure of asymmetry for each image is calculated. Since the facial expression is neutral in all three images, we would wish the measure of asymmetry to be the same in all three images. However, the left-most image is recorded as the most asymmetric (with a score of 31.57) and the right-most image is recorded as the second most asymmetric (with a score of 24.20). The score for the middle image - which represents the actual degree of asymmetry - is 19.19. The reason why our measure is affected by lateral rotation is that this kind of rotation makes the distance between a point on the face and the camera dependent on the point's x-coordinate (if we assume that the x-axis is perpendicular to the axis of symmetry), and this in turn means that features appear larger or smaller according to the side of the face they lie on and their distance from the axis of symmetry.  

In response to this limitation, videos must be filmed with minimal lateral rotation of the head, and with the head pointing directly towards the camera. To eliminate pose variation, the camera was attached to the head so that its position relative to the head was held constant. Unfortunately, it is not possible to have the camera perfectly straight on, as it may always be misaligned by a matter of millimetres. However, it is possible to have a constant alignment using this technique, and, since we are most interested in relative changes of asymmetry as emotional expression changes, rather than an absolute measure of asymmetry, constant alignment is more important than perfect alignment.

We undertook some research into overcoming this limitation by attempting to manipulate the image of a laterally rotated head to eliminate the effect of the rotation on our measure of asymmetry. To do this we used OpenCV's FindHomography and WarpPerspective functions. Imagine that the face can be represented by a plane. Lateral rotation of the face then amounts to lateral rotation of this plane. The FindHomography function is used to compute the homography matrix that represents the rotation of the plane. To use the function, we need to provide it with a set of coordinates for points on the plane prior to the rotation, and with a set of coordinates for the corresponding points after the rotation. In the case of rotation of the face, we choose a set of points on the face that are closest to lying in a plane. The function needs a minimum of 4 points, but more points are more likely to produce a good result.

Once we have computed the homography matrix we can use the WarpPerspective function to apply the perspective transform represented by the matrix to an image. The WarpPerspective function takes the homography matrix as input, along with the source image and the destination image. If we wish to attempt to eliminate the effect of the rotation on the face then we should use the \emph{inverse} of the homography matrix rather than the homography matrix itself.

\begin{figure}[htp]
\centering
\includegraphics[scale=0.44]{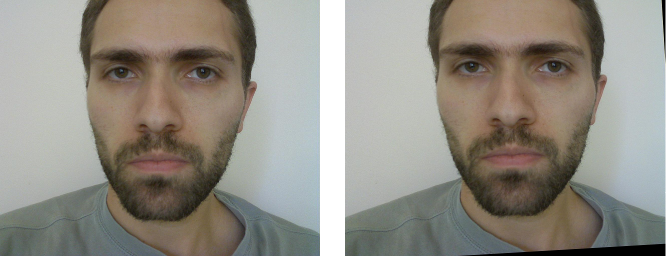}
\caption{The left image shows the slightly rotated head before the WarpPerspective function has been applied and the right image shows the image after the function has been applied. Asymmetry only slightly decreases; in the left image it is 31.57 and in the right image it is 30.15.}\label{fig:homography}
\end{figure}

To test the approach we calculated the homography matrix for a set of points on the middle image in Figure~\ref{fig:posevariation} and the corresponding set of points for the left-most image in the figure. This provided the homography matrix that estimated the transformation taking the middle image to the left-most image. We then inverted the matrix so that we had the transformation taking the left-most image to the middle image; in other words, the transformation to neutralise the rotation we see in the left-most image. Figure~\ref{fig:homography} shows the left-most image on the left - before we apply the inverted homography matrix - and then the image after we apply the inverted homography matrix (using the WarpPerspective function). Whilst it appears that the face has been somewhat ``rotated'' back towards the camera, calculating the asymmetry for both of the images in the figure reveals that the level of asymmetry is only negligibly reduced. If the method had performed as desired, the level of asymmetry after the inverted homography matrix had been applied would be significantly lower than before the application, and would be close to the measure recorded for the central image in Figure~\ref{fig:posevariation}. 

Calculating the homography matrix by using points on the face will always be an approximation because the human face is not a planar surface. However, it was hoped that by experimenting with different selections of points we could find an approximation good enough to allow our measure of asymmetry to be relatively unaffected by lateral rotation. We were unable to achieve this and decided instead to rely on fixed camera alignment by attaching the camera to the subject's head. A possible avenue of future research is to enhance our measure of asymmetry so that it is capable of coping with a range of pose variation. 
\\

This completes Chapter~\ref{sec:chapterthree}. In the next chapter we use Measure One to investigate the relationship between facial asymmetry and happiness.

\chapter{Applications of the measure}\label{sec:chapfour}

\section{About this chapter}

In the previous chapter we developed two measures of asymmetry, compared them, and decided that the first - which relies only on shape data - was preferable. We also described the measure's two main limitations; the first is that it is only as good as the fitting and the second is that lateral pose variation will affect the measure. However, if these two limitations are negated by ensuring that the fitting is good and that pose is constant we have an automatic frame-by-frame measure of facial asymmetry. Once the model is trained on a video it can calculate the level of asymmetry in any frame without relying on human input to annotate the frame or make measurements (this is in contrast to many early studies of asymmetry and emotion where humans were needed to measure the asymmetry in every image individually). This means that complete videos can be analysed and large datasets collected. In this chapter we collect some data and look at ways we can analyse them. 

\section{Facial asymmetry and happiness}

Due to time constraints, a single emotion was selected, and the relation between that emotion and facial asymmetry was studied in depth. It was decided that it would be more interesting to look at one emotion in detail rather than look at several emotions superficially. The emotion we selected was happiness; the reason was that it is easier to naturally elicit than, for example, fear or sadness. The work done on happiness could be generalised to other emotions providing we have videos of subjects experiencing those emotions, with constant pose and a good fitting. 

\subsection{Fitting}\label{sec:happyfitting}

\begin{figure}[htp]
\centering
\includegraphics[scale=0.3]{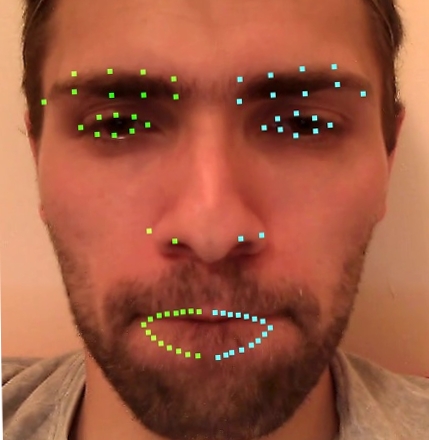}
\caption{The image shows an example of a frame where the shape data poorly fit the face because not enough training images have been used. We later added this frame to the training set so that the model was able to correctly fit faces where the lips were concealed (such as in this frame).}\label{fig:poorfitting}
\end{figure}

A ten-minute video was taken of the author viewing a comedy program on his computer. This elicited plenty of smiling, smirking and laughing and the video was devoid of negative emotions such anxiety or sadness. To train the model we first selected ten frames using a polyline algorithm (we explain how selection with a polyline algorithm works in Section~\ref{sec:training}). The model was then built and fitted to the video. To establish the quality of the fitting, the video was watched back with the shape data overlaying the video. Frames where the shape data were not good were noted (see Figure~\ref{fig:poorfitting} for an example of a frame with poor fitting). A selection of these frames was added to the training set and the model was retrained and again fitted to the video. This process was repeated until it was decided that the fit was good enough for analysis of the video. This point was reached when the training set contained 25 training images. At this point we randomly sampled 100 frames from the video to evaluate the quality of the fitting by eye. 93\% were fitted perfectly and the other 7\% were fitted very well (i.e., they had no more than 2 points out of place, and these points were out of place by no more than a few pixels). None of the frames in the sample of 100 were fitted poorly. 

\subsection{Analysing the video}\label{sec:analysis}

\begin{sidewaysfigure}[htp]
\centering
\includegraphics[scale=0.82]{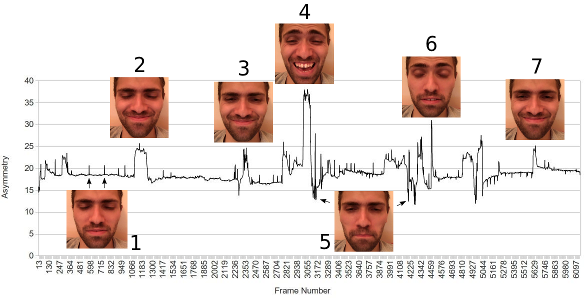}
\caption{Asymmetry plotted against frame number for a video of the author watching a comedy program.}\label{fig:happychart}
\end{sidewaysfigure}

The video was recorded at 15 frames per second. We selected a 6,139-frame segment (amounting to just under 7 minutes) where an interesting range of facial expressions were made. We then used the fitting and our measure of asymmetry (described in Section~\ref{sec:measureone}) to plot the graph shown in Figure~\ref{fig:happychart}. The graph shows facial asymmetry by frame for the 7-minute segment of video. Seven frames have been added to the graph at interesting points and we now discuss these.

\begin{itemize}
\item Image 1 illustrates that blinking has a small effect on our measure of asymmetry, increasing asymmetry slightly. Inspection revealed that this was due to the corner of the subject's left eye lowering slightly more than the corner of his right eye during blinking. Investigation of individual frames revealed that many of the small spikes on the chart were due to blinking.
\item Image 2 marks the point where asymmetry first exceeds a value of 25. This was the result of a broad smile. Inspection revealed that the subject's smile was more asymmetric than his neutral face because his mouth pulled slightly more to the left side of his face than the right during smiling, and his right eye shrunk more than his life eye. Further peaks of around 25 (see images 3 and 7 on the graph) were also caused by broad smiles.
\item Image 4 marks the point of greatest facial asymmetry caused by a strong laugh. Inspection of the laugh revealed the points of asymmetry; the subject's left eyebrow arched more than his right and the left side of the lower lip had stronger curvature than the right side. As with his smile, the subject's right eye closed a little more than his left.
\item Interestingly, the three points where facial asymmetry dropped below 15 (and where the face was therefore most symmetrical) were due to pursing of the lips, which compacted and rounded the lips, as illustrated in Image 5. 
\item Raising eyebrows typically increased asymmetry due to the left eyebrow raising and arching more than the right. Image 6 is an example of this.
\end{itemize}

\section{Measuring left-sided, right-sided and overall facial movement}

\begin{figure}[htp]
\centering
\includegraphics[scale=0.22]{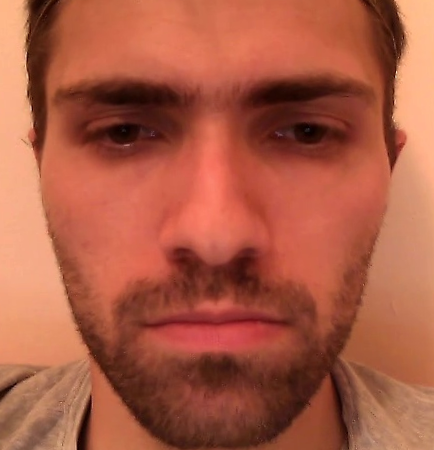}
\caption{The image shows the frame designated as the neutral frame of the video. Overall movement (and left-sided and right-sided movement) was measured as movement from this frame.}\label{fig:neutralframe}
\end{figure}

Figure~\ref{fig:happychart} suggests that smiles and laughter are associated with increased facial asymmetry. This can be extended to a more general hypothesis that the degree of asymmetry increases as the strength of the ``happiness expression'' increases (for example, a slight smile is more asymmetric than a neutral face; a strong smile is more asymmetric than a slight smile; and laughter is more asymmetric than a strong smile). To further investigate this hypothesis, we needed a measure of the degree of happiness expressed. To achieve this, we designated a frame in the opening part of the video as the ``neutral frame'' (the frame is shown in Figure~\ref{fig:neutralframe}). We decided to measure ``degree of happiness'' as overall movement away from this neutral face. This approach is only valid if the video only contains either neutral or happy expressions. If the video also contained - for example - fearful expressions, then we could not simply identify the level of movement from the neutral face with the degree of happiness expressed. 

To measure movement from the neutral face we measured left-sided and right-sided movement from the neutral face separately and used the average as a measure of overall movement. The reason that we did this was to connect our work with topics in neuropsychology discussed in Chapter 2. Recall that the right hemisphere hypothesis conjectures that the right side of the cerebral hemisphere is responsible for processing both positive and negative emotions whereas the valence hypothesis conjectures that the right side is only responsible for negative emotions and the left side is responsible for positive emotions. Recall also that the right hemisphere hypothesis predicts greater movement on the left side of the face than on the right during the expression of emotions whereas the valence hypothesis predicts greater movement on the left side for negative emotions and greater movement on the right for positive emotions. To see if our video confirms either hypothesis, we would need measures of left-sided movement and right-sided movement from the neutral face. In the next section we explain how we developed these measures.

\subsection{Developing the measures}\label{sec:lrmeasure}

\begin{figure}[htp]
\centering
\includegraphics[scale=0.39]{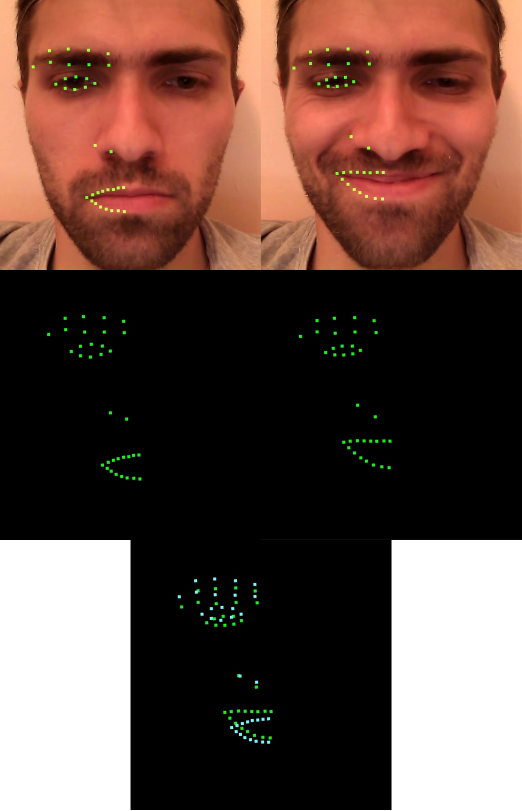}
\caption{The figure illustrates how we calculate left-sided movement between the neutral frame (on the left of the top row) and another frame. The result of Procrustes analysis to the two sets of points is shown on the third row. The Procrustes distance - which is our measure of degree of movement from the neutral frame - is 74.94.}\label{fig:sidemovement}
\end{figure}

Building a measure of left-sided and right-sided movement on top of our measure of asymmetry was relatively straightforward. Shape data for the fitted neutral frame are stored. These shape data are separated into two sets: data for the left side of the face and data for the right side of the face. Given a frame, to measure left-sided movement from neutrality, we apply Procrustes analysis to the shape data for the left side of the face to determine the translation and rotation that minimises the Procrustes distance to the shape data for the left side of the neutral face. The minimised Procrustes distance is our measure of left-sided movement from the neutral frame. To measure right-sided movement, we do the same for the right side of the face. Overall movement from the neutral face is simply defined as the average of left-sided and right-sided movement. Figure~\ref{fig:sidemovement} illustrates the calculation of left-sided movement for a frame of the video where the subject is smiling. The top row shows the neutral frame on the left and the frame with the subject smiling on the right. We want to calculate the degree of shape difference between these two frames. On the second row we see the left-sided shape data for each frame. And on the third row we see these points after Procrustes analysis has been used to minimise their Procrustes distance using only translation and rotation. The Procrustes distance - which is our measure of degree of movement from the neutral frame - is 74.94. 

\subsection{Charting left and right-sided movement}\label{sec:lrmov}

\begin{figure}[htp]
\centering
\includegraphics[scale=0.36]{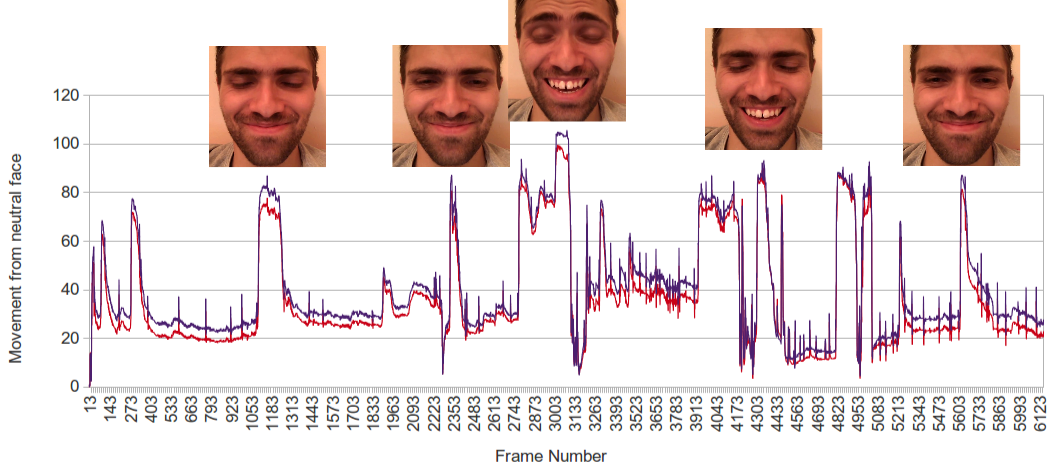}
\caption{The chart shows facial movement from the neutral face on the y-axis against frame number on the x-axis for a video of the subject watching a comedy program. The neutral face was defined as the shape data for the face in frame 0 (where the subject's face was expressionless). This means that the graph starts at the point (0, 0). The magenta line records movement of the left side of the subject's face and the red line records movement of the right side of the subject's face (where left and right are defined from the subject's perspective).}\label{fig:leftright}
\end{figure}

With our measures of left-sided and right-sided movement we plotted the graph in Figure~\ref{fig:leftright} for the same video that was used for Figure~\ref{fig:happychart}. The first frame (i.e., frame 0) was designated as the neutral face because it contained an expressionless face. The y-axis records movement from this neutral frame (which means that the graph starts at (x, y) = (0, 0)). Magenta is used to plot movement of the left side of the subject's face and red is used to plot movement of the right side of the subject's face (where left and right are from the subject's perspective). We have annotated five peaks on the graph with the frames that correspond to those peaks. Reassuringly, peaks coincide with frames where the face expresses a happy expression. Moreover, the only point where movement exceeds 100 coincides with the point where the happy expression is strongest and the subject is laughing. This makes sense because the subject's laughter involves not only movement of his lips and nose, but also his eyebrows are raised. We can also see that the magenta line appears to be higher than the red line for nearly all frames. The data confirm this: movement of the left side of the subject's face exceeded movement of the right side in 97.3\% of the video's frames.

\begin{figure}[htp]
\centering
\includegraphics[scale=0.74]{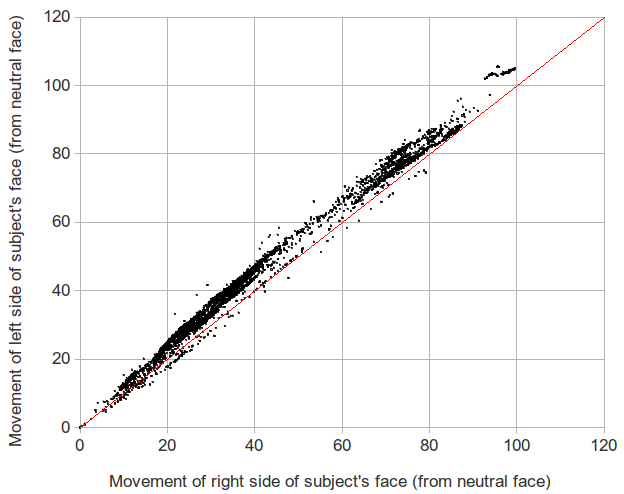}
\caption{The figure shows a scatter plot of left-sided movement on the y-axis and right-sided movement on the x-axis. The line y = x is drawn in red.}\label{fig:lrscatter}
\end{figure}

Figure~\ref{fig:lrscatter} is a scatter graph plotting left-sided movement against right-sided movement for the video (each point of the graph represents a frame). The straight line y = x is drawn in red to show that most points lie above the line which means that the majority of frames exhibit more movement on the subject's left side than on his right side. We also see that movement on the left side is closely correlated to movement on the right side. The Pearson correlation coefficient was measured to be 0.996 which shows a very strong dependency (a coefficient of 1 means a perfect correlation). What we can conclude, then, about the relation between left and right-sided movement (for this particular subject, expressing the particular emotion ``happiness'') is that left and right movement happen simultaneously but that the degree of left movement nearly always exceeds the degree of right movement. This latter fact counts as evidence against the valence hypothesis and in favour of the right hemisphere hypothesis (because the valence hypothesis predicts greater right-sided movement for positive emotions whereas the right hemisphere hypothesis predicts greater left-sided movement for positive emotions).  

\subsection{Charting asymmetry against overall movement}\label{sec:asymov}

\begin{figure}[htp]
\centering
\includegraphics[scale=0.5]{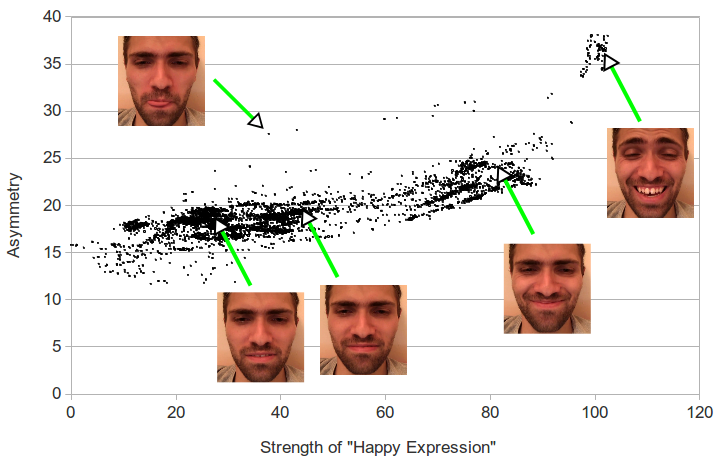}
\caption{The figure shows a scatter graph displaying strength of happiness expression on the x-axis and asymmetry on the y-axis. The graph shows a clearly positive correlation (the Pearson coefficient is 0.77), with an increase in the strength of happiness typically being accompanied by an increase in asymmetry.}\label{fig:happyasymmetry}
\end{figure}

One of the reasons for putting measures of left and right-sided movement on the video was that they can be used to get a measure of overall movement (by averaging left and right-sided movement) and we can then see if overall movement is correlated to asymmetry. Figure~\ref{fig:happyasymmetry} is a scatter graph showing overall movement from the neutral face on the x-axis and asymmetry on the y-axis. Remember that we are using overall movement as our measure of the strength of happiness expression. The graph shows a clearly positive correlation (the Pearson coefficient is 0.77), with an increase in the strength of happiness typically being accompanied by an increase in asymmetry. We can see that the relationship is not linear; it seems that earlier increases in the strength of expression (for example, increases from 20 to 40) make less of a difference to asymmetry than later increases (for example, increases from 80 to 100). We have annotated the plot with some frames from the video. Values of x between 20 and 60 are associated with slight smiles or smirks; asymmetry for these faces is not much greater than asymmetry of the neutral face (which has asymmetry of 15.81). Values of x between 70 and 90 are associated with broader smiles which increase the level asymmetry to around 20 to 25. The top-right small cluster of points marks the period of the video when the subject was laughing, which was associated with the greatest asymmetry. There are a few points that lie outside the main body of points. One of these outliers is annotated at the top-left of the graph. This is a frame where the asymmetry over strength of expression ratio is higher than for most frames. The reason for this is that the subject's tongue is distorting the shape of the mouth, increasing asymmetry, but the face is still fairly close to the neutral face. 

\section{Analysing a second subject}\label{sec:secondsub}

To compare the results obtained thus far, a 10-minute video of another subject watching a comedy program was recorded. The second subject was male, one year younger than the first subject and - like the first subject - right-handed. As with the first video, we selected a segment of the video that captured a range of emotional expression. This segment contained 5,817 frames and was six and a half minutes in length. We trained an AAM for the video until we achieved fitting performance equivalent to the performance for the first video. We then ran the asymmetry measure on the segment to return 5,817 data points. For each data point we had a measure of asymmetry, a measure of left-sided movement from the neutral frame (where the neutral frame was the first frame, where the subject's face was expressionless) and a measure of right-sided movement from the neutral frame. We were then able to plot scatter graphs analogous to Figures~\ref{fig:lrscatter} and~\ref{fig:happyasymmetry} for this second subject.

\begin{figure}[htp]
\centering
\includegraphics[scale=0.55]{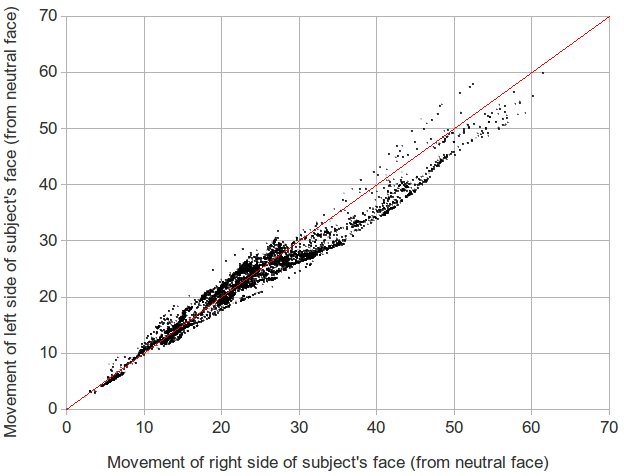}
\caption{The figure is a scatter graph for the second subject of left-sided movement (from the neutral face) on the y-axis and right-sided movement (from the neutral face) on the x-axis. The line y=x is drawn in red.}\label{fig:edizlr}
\end{figure}

Figure~\ref{fig:edizlr} is a scatter graph showing left-sided movement from the neutral face on the y-axis, and right-sided movement from the neutral face on the x-axis. Comparing this graph to the graph in Figure~\ref{fig:lrscatter} there are two important differences. The first is that movement for both sides of the face ranges from values of 0 to just over 60, whereas for the first video, movement ranged from values of 0 to over a 100. This suggested that the expressions of happiness elicited from the second subject were not as strong as the first subject. Watching the video showed that this was indeed the case. Either through inhibition, or because the comedy program was not funny enough, whilst the subject smiles and grins at points, the subject does not laugh at any point in the video. The second difference to Figure~\ref{fig:lrscatter} is that the right side of the face shows dominance over the left, at least for strong expressions (in contrast, for the first subject, a clear left-sided dominance was found, with 97.3\% of frames showing greater left-sided movement than right-sided movement). 85.3\% of frames showed average movement of less than 30. Of these frames, 48.4\% of frames showed greater left-sided movement, and 51.6\% showed greater right-sided movement. However, for the frames that showed average movement greater than 30, only 7\% showed greater left-sided movement and 93\% showed greater right-sided movement. Our findings for the second subject provide support to the valence hypothesis, which predicts greater right-sided movement for expression of positive emotions.

\begin{figure}[htp]
\centering
\includegraphics[scale=0.47]{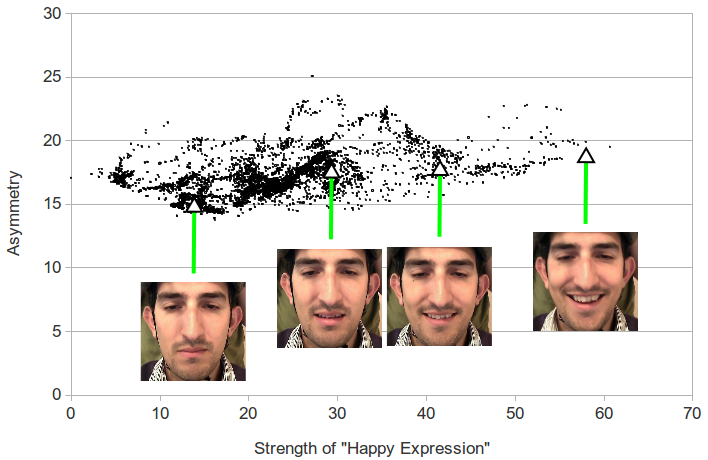}
\caption{The figure shows a scatter graph of facial asymmetry on the y-axis and strength of the ``happiness expression'' on the x-axis for our second subject.}\label{fig:edizhappy}
\end{figure}

We can also plot a scatter graph of magnitude of asymmetry against strength of ``happiness expression'' for the second subject, analogous to Figure~\ref{fig:happyasymmetry}. This is shown in Figure~\ref{fig:edizhappy}. We have added 4 images from the video to annotate the graph. The left-most image shows the subject with a neutral face and the degree of asymmetry is low. The next image (from the left) shows the subject with some slight movement away from the neutral face (movement is just under 30) and asymmetry elevated slightly, but not by much. The third image shows the subject with a slight smile. As with the second image, asymmetry is elevated, but only slightly (compare, for instance, the range of asymmetry values for the first subject in Figure~\ref{fig:happyasymmetry} - asymmetry gets close to 40 when the subject is laughing). The right-most image shows one of the most expressive frames of the video (i.e., a frame with high movement from the neutral face). The subject is seen to be smiling and the asymmetry value is around 20. 

It is hard to infer from Figure~\ref{fig:edizhappy} that asymmetry increases as the strength of expression increases for our second subject. Asymmetry is seen to increase slightly as the strength of expression increases, but the trend is not as clear as for our first subject. It is worth noting that strength of expression ranges from 0 to 60 for the second subject in contrast to the first subject where it exceeded 100 when the subject was laughing strongly. If we were able to record a second video of the second subject that contained strong laughter, then we could see if stronger expressions of happiness are associated with larger increases in asymmetry, as for our first subject. 

This concludes the chapter. In the next and final chapter we review our work and subject it to a critical evaluation.

\chapter{Conclusion}\label{sec:conclusion}
In this chapter we summarise our project and report; submit the former to a critical evaluation; and suggest possible future directions for research.
\section{Summary of project and report}
Our project was to:
\begin{itemize}
\item develop an automatic frame-by-frame measure of facial asymmetry in videos of faces that improved on previous measures (automatic or otherwise), and
\item use the measure to analyse the relationship between facial asymmetry and emotional expression, and connect our findings with previous research of the relationship
\end{itemize}
Our report was divided into three chapters (excluding the introductory chapter and this chapter). In Chapter~\ref{sec:chaptwo} we provided the reader with the motivation for our work and its theoretical basis. We began by showing why anyone should want to measure facial asymmetry. Psychologists have long been interested in the asymmetry of faces in static, neutral poses (and have connected the degree of asymmetry to attractiveness, health and personality type) and neuropsychologists have investigated the dynamics of asymmetry during emotional expression, to understand emotional processing in the brain. We argued that there was an opportunity to improve on previous measures of facial asymmetry and provide neuropsychologists with new tools to collect data concerning the relationship between asymmetry and emotional expression. Earlier measures relied on human input (to measure any given image) and were therefore time-consuming, limiting the quantity of data that could be collected for a study. More recent measures (based on techniques from computer vision) that could theoretically collect larger datasets were found to be limited in number. Active Shape Models and Active Appearance Models were selected as the basis of a new, automatic measure of facial asymmetry, without the limitations of those previously discussed, and the theory underlying them was presented in detail.

In Chapter~\ref{sec:chapterthree} we thoroughly described the work that was done to develop two measures of facial asymmetry. After selecting an AAM library (and justifying its selection) we established the library's capability by training it on a test video and evaluating its fitting performance. We then used the library as the basis for the development of the two measures. The first measure used shape data to measure asymmetry; shape data for each side of the face were collected and Procrustes analysis was used to minimise the Procrustes distance between the shape data for the right side and the shape data for the left side (through reflection, rotation and translation). The minimised distance was taken as the degree of asymmetry. The second measure used both shape data and texture data. The former were used to calculate the axis of symmetry for the face. Texture was then reflected across the axis, and asymmetry was defined as the mean absolute difference between corresponding pixel values for the reflected and unreflected sides of the face. The two measures were critically compared, and the first was deemed to be preferable for several reasons. Two limitations of the first measure were identified: first, that it was only as good as the quality of the shape data provided by the fitting; second, that it could be affected by lateral head rotation, which therefore had to be eliminated by - for example - attaching the camera to the subject's head.

Chapter~\ref{sec:chapfour} described applications of the measure and analysed data that were collected. Due to time constraints, a single emotion - happiness - was selected for analysis. A seven-minute segment of video of a subject viewing a comedy program was analysed in detail. Several interesting results were found. Analysis of a graph of asymmetry against frame number (Figure~\ref{fig:happychart}) highlighted frames of elevated asymmetry in the video; analysis of the shape data for these frames identified surprising causes of asymmetry (such as the subject raising his eyebrows). Measures of left and right facial movement (relative to a neutral frame) were developed and were used to plot a scatter graph of left-sided movement against right-sided movement, which confirmed a clear bias in favour of left-sided movement for the subject (Figure~\ref{fig:lrscatter}). In contrast, results for a second subject (Figure~\ref{fig:edizlr}) indicated a bias in favour of right-sided movement (at least for stronger expressions). The results were connected with our research into topics in neuropsychology; the results for the first subject supported the right hemisphere hypothesis and the results for the second subject supported the valence hypothesis. Scatter graphs of asymmetry against the strength of the ``happiness expression'' were also plotted. A positive, but non-linear, relationship between the variables was observed (Figure~\ref{fig:happyasymmetry}) for the first subject. The results for the second subject (Figure~\ref{fig:edizhappy}) were inconclusive.  

\section{Critical evaluation of work}\label{sec:evaluation}

To perform a critical evaluation of the project we will consider its strengths and weaknesses in turn.

\subsection{Strengths}\label{sec:strengths}

The first aim of the project was to develop an automatic frame-by-frame measure of facial asymmetry in videos of faces that improved on previous measures. Two measures of asymmetry were developed, and the first was preferred, for reasons discussed in Section~\ref{sec:comparison}. We believe that the development of the first measure satisfies the first aim of our project. Given a video, and an ASM trained for that video, our measure returns a value of asymmetry for every frame, without human involvement (i.e., is automatic). To be sure, training of the ASM involves human input, as training images need to be annotated by hand. However, we have found that a high-quality model for a ten-minute video can be trained from as few as 25 images (Section~\ref{sec:happyfitting}), where each image takes 2 to 3 minutes to annotate. A ten-minute video (at 15 frames per second) yields 9,000 frames of asymmetry data. 

Our measure improves on earlier measures - that relied wholly on human measurement - because it can collect more data for analysis. (To measure the asymmetry of 9,000 frames, by hand, would likely take several days.) But it also improves on other automatic measures of asymmetry in the literature. To our knowledge, no one has previously used ASMs or AAMs to measure facial asymmetry; nor has anyone previously developed an automatic Procrustes-based measure. As we saw in Section~\ref{sec:litrev2}, when we looked at other automatic measures, \cite{d09} relied on an entropy-based measure (that looked at average changes in pixel values across each side of the face) and \cite{necy04} used a 3-dimensional range finder to detect movement. Each measure had self-confessed limitations. Desai assumed that ``changes in the surface lighting of the face reflect movement'' and his measure was therefore affected by lighting variation, and Nicholls and colleagues stated that they could only measure movement perpendicular to the surface of the face. However, the main limitation of both measures, in our view, was that they treated every pixel or coordinate of the face equally, and were unable to ignore asymmetries that, for whatever reason, we may wish to exclude from our measure (such as movement of the eyes - which cannot be ignored by Desai - or uneven raising of the eyebrows - which cannot be ignored by Nicholls). A Procrustes-based measure, such as ours, is based only on the chosen set of landmarks, and so we can choose to ignore any feature of the face. Furthermore, if we want to increase the weighting of a feature in the measure, we need only add extra points to it. 

It is for these reasons that we believe that our novel measure, as an automatic Procrustes-based measure, improves on earlier measures. First, it improves on earlier measures which rely on human measurement because it can collect more data. And second, it improves on measures that are not Procrustes-based, because of its ability to specify and weight the features of the face that count towards the measure.

The second aim of the project was to use our measure to investigate the relationship between facial asymmetry and emotional expression, and connect our findings with previous research of the relationship. We pursued this aim in Chapter~\ref{sec:chapfour}. A seven-minute segment of video (of just over 6,000 frames) of the author watching a comedy program was analysed in detail. Novel Procrustes-based measures of left and right-movement from a neutral frame were developed (it should be noted that these measures share the same strengths as our measure of asymmetry; namely, they are automatic and allow the user to choose which features of the face are tracked). We discovered a clear bias in favour of left-sided facial movement for the subject during expressions of happiness, and we connected this result to work in the literature (as support for the right hemisphere hypothesis). For a second subject, we discovered a bias in favour of right-sided facial movement during stronger expressions of happiness (and counted this as support for the valence hypothesis). We used our measures of left and right-movement to define a measure of the strength of the ``happiness expression''. To our knowledge, no one in the literature has previously investigated the relationship between the strength of an emotional expression and the degree of its asymmetry (studies have instead focussed on the direction of asymmetry - i.e., which side of the face moves more - rather than on its magnitude). Figure~\ref{fig:happyasymmetry} illustrates the discovery of an interesting non-linear but positive relationship between the strength of the ``happiness expression'' and its degree of asymmetry for our first subject. The data for our second subject do not rule out the same relationship. On the basis of our analysis, we were able to form a hypothesis worthy of future research: that magnitude of asymmetry increases with strength of emotional expression, with stronger expressions especially associated with stronger asymmetry.  

Although these results are limited to two subjects and a single emotion, similar data could be collected for further subjects and emotions. We discuss this further in Section~\ref{sec:furtherwork}. 

\subsection{Weaknesses and possible improvements}

We believe that there are three genuine weaknesses of our project, and one aspect of the project that could be perceived as a weakness, but that is not a genuine weakness. Two of the three genuine weaknesses apply to our measure, and the other applies to our analysis of the relationship between asymmetry and emotional expression. 

The aspect of the project that could be perceived as a weakness, but that is not a weakness, is that time was invested into developing a second measure of asymmetry (Section~\ref{sec:measuretwo}), but the measure was eventually discarded and the first measure was preferred. And further, if time had not been invested into the second measure, there would have been more time to collect and analyse data for additional subjects. There are two reasons why this is not a genuine weakness of the project. The first reason is that development of the second measure was necessary to realise its weaknesses, and to appreciate the strengths of the first measure. The chief advantage of the first measure over the second measure is its flexibility; because the measure is based only on the chosen set of landmarks, the user can choose to ignore any feature of the face that they wish to, and focus on certain features of the face. We saw in Section~\ref{sec:strengths} that this is also the chief advantage of the first measure over other recent measures in the literature. It was by developing the second measure that we were able to perceive this advantage. The second reason why developing the second measure was not a weakness of the project is that even if we had collected data for additional subjects, we would - at most - have had time to collect data for 5 - 10 subjects, and this sample size would not have been large enough to draw generalisable conclusions worthy of publication, and so we feel that focussing on the development of the measures was more valuable.

The first genuine weakness of our measure is that it is affected by lateral rotation of the subject's head, which means that our measure can only reliably analyse videos where lateral rotation is absent or negligible (see Section~\ref{sec:posevariation}). Since lateral rotation of a subject's head is natural during speech and emotional expression, our measure can only be performed on videos where the subject has been asked not to turn their head sideways, or where the camera has been attached to the subject's head to eliminate rotation. This means that our measure can only be used for specially recorded videos and cannot be applied to - for example - corroborate asymmetry measures in earlier studies. However, whilst this is a limitation of our measure, we do not believe that it is a serious one. First, new studies into asymmetry and emotional expression will most likely want to collect new data, and attaching the camera to the subject's head is nothing more than an inconvenience. And second, the limitation provides an opportunity to undertake research into building on our measure so that it can cope with lateral head rotation. 

The second genuine weakness applying to the measure is that training a model for a subject involves manually annotating images, which typically takes 2-3 minutes per image, and therefore limits the amount of data that can be collected for a study. It should be noted that this is not a weakness of the Procrustes-based approach to measuring asymmetry or of our implementation of it. Rather it is a consequence of the fact that the AAM library that was selected required training and did not provide a pre-trained model. In addition, we should say that we never experimented with a model trained on more than 25 training images. It may be that if we were conducting a large study of facial asymmetry and emotional expression with, for instance, 100 subjects, we could train the model on a large set of training images (i.e., several hundred) drawn from across the study, and the model would fit all subjects well and generalise to new subjects. However, the initial training would still involve a substantial time investment, and so a possible improvement of our measure is to integrate it with a method of fitting that does not require training. In communication with Alexander Davies (PhD student at the University of Bristol), it has recently come to our attention that a face tracker due to Jason Saragih (based on \cite{slc11}) performs well on a range of videos (accurately locating facial features) and comes with a pre-trained model, and requires no additional training. It may be that combining this face tracker and our Procrustes-based measure of asymmetry could yield a fully automatic measure of asymmetry. 

The final weakness of our project, that applies to our analysis of the relationship between asymmetry and emotional expression, is that we only collected data for two subjects and a single emotion (happiness). The reason for this was simply that developing and assessing the measures, and producing this report, took the majority of the time available for the project. However, even though we were unable to analyse data for more than two subjects, we believe that we have prepared the ground for further research worthy of publication. In particular, our analysis of the correlation between the magnitude of facial asymmetry and strength of emotional expression is original (i.e., Figures~\ref{fig:happyasymmetry} and~\ref{fig:edizhappy}) and allows us to postulate a hypothesis worthy of further investigation: that magnitude of asymmetry increases with the strength of emotional expression where strong expressions (such as laughter) are especially associated with strong asymmetry. Furthermore, our measure could be used for further research into determining whether positive emotions are associated with greater movement of the left side of the face or of the right side (or of neither side). This remains an open question in the literature.

\section{Future directions}\label{sec:furtherwork}

We end the report by suggesting three interesting and fruitful ways to build on the research undertaken. 

The first way is it to attempt to develop the measure so that it can cope with lateral head rotation. If this could be accomplished, new videos for analysis could be recorded without the camera attached to the subject's head, and videos from previous studies on emotional expression could be analysed. This is probably the most difficult of the three suggested paths for future research. As we saw in Section~\ref{sec:posevariation}, our attempt to handle lateral rotation using OpenCV's homography functions was unsuccessful. What is needed, it seems, is a way of estimating the depth of each point on the face (i.e., distance from the camera lens). This information could then be used to more accurately estimate the degree of lateral rotation of the head, and factor this degree into the measure of asymmetry. However, we remain sceptical at this point that a 2-dimensional measure could be developed that was wholly unaffected by head rotation, and further research is certainly needed.

The second way to build on our research is to focus on increasing the degree of automation of our measure, by integrating it with a face tracker that does not require manual training. Alexander Davies has recently started using a face tracker due to Jason Saragih (based on \cite{slc11}) that does not need training and - according to Alexander - can effectively fit new faces. A fruitful path for future research would be to integrate this face tracker with our Procrustes-based measure; the tracker would be used to locate facial landmarks, and our measure would be used to calculate the degree of asymmetry from the landmarks. The end result would be a measure of facial asymmetry that is fully automatic. The benefit over our measure would be that a new study of facial asymmetry - with, for instance, 100 subjects - would not require a laborious training stage involving several hundred training images and taking several hours. 

The final way of building on our research is to use the developed measure for a substantial study (of as many subjects as possible) that seeks to answer the following:
\begin{itemize}
\item Is emotional expression, for positive emotions, associated with greater movement on the left side of the face (as the right hemisphere hypothesis predicts), or the right side of the face (as the valence hypothesis predicts)? We have found greater movement on the left side for one subject, but greater movement on the right side for another. Time permitting, we would have liked to have collected data for additional subjects.
\item Is strength of emotional expression positively correlated with degree of facial asymmetry. For happiness, and for one subject, we have found that it is. For a second subject, the data are inconclusive. Our measure provides the opportunity to collect data for further subjects, and determine the correlation.
\end{itemize}
This concludes Chapter 5. We hope to have convinced the reader that our research has made valuable contributions to the problem of measuring facial asymmetry in videos, and performed original analysis into the relationship between facial asymmetry and emotional expression, that is interesting in its own right, and that can teach us about emotional processing in the brain. We hope further that we, or other researchers, have the opportunity to build on our research by following one of the paths described above.
\vspace{30pt}
\begin{center}
------------------------------------
\end{center}

\bibliographystyle{apalike}
\bibliography{simple}
\appendix

\chapter{ASM and AAM software implementations}\label{sec:appendixone}
The table below displays the URLs associated with the ASM and AAM software implementations presented in Section~\ref{sec:softwareimp}.
\begin{table}[htcp]\footnotesize
\centering
\begin{tabular}{|p{0.3cm}|p{1.5cm}|p{10.3cm}|}
\hline
& Author & URL \\ \hline
1 & Ghassan Hamarneh & \url{http://www.cs.sfu.ca/~hamarneh/software/asm/index.html}\\ \hline
2 & Tim Cootes & \url{http://personalpages.manchester.ac.uk/staff/timothy.f.cootes/software/am_tools_doc/index.html}\\ \hline
3 & Stephen Milborrow & \url{http://www.milbo.users.sonic.net/stasm/}\\ \hline
4 & Yao Wei & \url{http://code.google.com/p/asmlibrary/}\\ \hline
5 & Mikkel B. Stegmann & \url{http://www2.imm.dtu.dk/~aam/}\\ \hline
6 & George Papandreou & \url{http://cvsp.cs.ntua.gr/software/AAMtools/}\\ \hline
7 & Yao Wei & \url{http://code.google.com/p/aam-library/}\\ \hline
\end{tabular}
\end{table}
% put the links in the appendix
%http://www.cs.sfu.ca/~hamarneh/software/asm/index.html
%http://www.mathworks.com/matlabcentral/fileexchange/26706-active-shape-model-asm-and-active-appearance-model-aam
%http://personalpages.manchester.ac.uk/staff/timothy.f.cootes/software/am_tools_doc/index.html
%http://www.milbo.users.sonic.net/stasm/
%http://code.google.com/p/asmlibrary/
%http://www2.imm.dtu.dk/~aam/
%http://cvsp.cs.ntua.gr/software/AAMtools/
%http://code.google.com/p/aam-library/
\chapter{A theorem about reflection}\label{sec:reflection}
The following theorem was required in Section~\ref{sec:measuretwo}. Its proof is provided here.
\newline
\\
\emph{Theorem.}
Reflection of the point $(x', y')$ in the line $y = mx$ takes the point to:
\[(\frac{2my' - (m^2 - 1)x'}{1 + m^2}, \frac{(m^2 - 1)y' + 2mx'}{1 + m^2})\]

\begin{proof}
Let $L_{1}$ be the line $y = mx$. Let $L_{2}$ be the line perpendicular to $L_{1}$ that passes through $(x', y')$. Then $L_{2}$ has the form $y = \frac{-x}{m} + c$. Let $(x'', y'')$ be the intersection of $L_{1}$ and $L_{2}$. Since $(x', y')$ lies on $L_{2}$ and $(x'', y'')$ lies on $L_{1}$ and $L_{2}$, we have the following:

\begin{equation}\label{eq:one}
y' = \frac{-x'}{m} + c
\end{equation}
\begin{equation}\label{eq:two}
y'' = \frac{-x''}{m} + c
\end{equation}
\begin{equation}\label{eq:three}
y'' = mx''
\end{equation}
We have three equations and three unknowns ($x'', y''$ and $c$). Combining Equations~\ref{eq:one} and~\ref{eq:two} yields $y' + \frac{x'}{m} = y'' + \frac{x''}{m}$. Substitute in Equation~\ref{eq:three} to find that $y' + \frac{x'}{m} = mx'' + \frac{x''}{m}$. Thus, $x'' = \frac{my' + x'}{1 + m^2}$ and $y'' = \frac{m(my' + x')}{1 + m^2}$.
\newline
\\
The coordinates of the reflected point are $(x''', y''')$ where $x''' = 2*x'' - x'$ and $y''' = 2*y'' - y'$. Thus, the reflected point is
\[(2*\frac{my' + x'}{1 + m^2} - x', 2*\frac{m(my' + x')}{1 + m^2} - y')\]
which, after some algebra, simplifies to the desired result.

\end{proof}
%http://www.sonoma.edu/users/w/wilsonst/papers/geometry/isometries/default.html
\chapter{Source code excerpt}
In this appendix we provide a source code excerpt. The code below implements the first measure (i.e., the measure described in Section~\ref{sec:measureone}). The code needs to be run with the AAM library and OpenCV.
\\
\newline
\lstset{basicstyle=\tiny, tabsize=2, breaklines=true, showspaces=false, showtabs=false, showstringspaces=false,
numberstyle=\tiny, language=C++, numbers=left}
\begin{lstlisting}[language=C++]
#include "AAM_IC.h"
#include "AAM_Basic.h"
#include "AAM_MovieAVI.h"
#include "VJfacedetect.h"
#include "global.h"
#include "gnuplot_i.h"
#include <math.h>

#define GREEN 1

using namespace std;

float getMedian(CvMat* pts);
void splitPoints(CvMat* pts_left, CvMat* pts_right, CvMat* pts, float median);
void pointsOnImage(IplImage* image, CvMat* pts_left, CvMat* pts_right, int frameno);
void printPts(CvMat* A);
void PrintMat(CvMat* A);
void pointsOnBlack(IplImage* image, CvMat* pts_left, CvMat* pts_right, char* windowname, int frameno);
void doProcrustes(CvMat* T, CvMat* translation, CvMat* A, CvMat* B);
void GetMean(CvMat* src, CvMat* dst);
void translate_to_origin(CvMat* pts);
void normalise(CvMat* pts);
void translate(CvMat* src, CvMat* translation, CvMat* dst);
float ProcrustesDistance(CvMat* one, CvMat* two, FILE* f, int frameno);
bool inFrames_to_save(int a);
void overlayPoints(IplImage* dst, IplImage* image, CvMat* points, int colour);

static void print_version()
{
	printf("\n\n"
		   "/**************************************************************/\n"
		   "/*     AAMLibrary -- A C++ open source for face alignment     */\n"
		   "/*  Copyright (c) 2008-2009 by Yao Wei, all rights reserved.  */\n"
		   "/*                    Contact: njustyw@gmail.com              */\n"
		   "/**************************************************************/\n"
		   "\n\n");
}

static void usage()
{
	printf("Usage: fit model_file cascade_file (image/video)_file\n");
	exit(0);
}

int main(int argc, char** argv)
{
	print_version();

	if(argc != 4)	usage();
	
	AAM_Pyramid model;
	model.ReadModel(argv[1]);
	VJfacedetect facedet;
	facedet.LoadCascade(argv[2]);
	char filename[100];
	strcpy(filename, argv[3]);	

	if(strstr(filename, ".avi"))
	{
		AAM_MovieAVI aviIn;
		AAM_Shape Shape;
		IplImage* image2 = 0;
		bool flag = false;
		
		FILE* pFile;
		pFile = fopen("results_asymm.txt", "w");
		fputs("frame number, asymmetry\n", pFile);

		FILE* fp;
		fp = fopen("point_distances.txt", "w");

		aviIn.Open(filename);
		
		for(int j = 0; j < aviIn.FrameCount(); j ++)
		{
			printf("Tracking frame %04i: ", j);

			IplImage* image = aviIn.ReadFrame(j);
			
			if(j == 0 || flag == false) 
			{
				flag = model.InitShapeFromDetBox(Shape, facedet, image);
				if(flag == false)
					continue;
			}
	
			model.Fit(image, Shape, 30, false);
			if(image2 == 0) image2 = cvCreateImage(cvGetSize(image), image->depth, image->nChannels);
			cvZero(image2);
			model.Draw(image2, Shape, 2);

			char filename[100];

			if (inFrames_to_save(j)) {
				sprintf(filename, "pointsImage%d.jpg", j);
				IplImage* imageallpts = cvCreateImage(cvGetSize(image), image->depth, image->nChannels);
				overlayPoints(imageallpts, image, points_all, GREEN);
				cvSaveImage(filename, imageallpts);
				cvReleaseImage(&imageallpts);
			}

			float median = getMedian(points_all);

			CvMat* pts_left = cvCreateMat((points_all->cols) / 4, 2, CV_64FC1);
			CvMat* pts_right = cvCreateMat((points_all->cols) / 4, 2, CV_64FC1);
			CvMat* pts_right_rotated = cvCreateMat((points_all->cols) / 4, 2, CV_64FC1);
			CvMat* pts_right_translated = cvCreateMat((points_all->cols) / 4, 2, CV_64FC1);

			splitPoints(pts_left, pts_right, points_all, median);

			pointsOnImage(image, pts_left, pts_right, j);

			pointsOnBlack(image, pts_left, pts_right, "pointsblack", j);

			CvMat* T = cvCreateMat(2, 2, CV_64FC1);
			CvMat* translation = cvCreateMat(1, 2, CV_64FC1);

			doProcrustes(T, translation, pts_left, pts_right);

			cvGEMM(pts_right, T, 1, NULL, 0, pts_right_rotated, 0);
			pointsOnBlack(image, pts_left, pts_right_rotated, "rotatedpoints", j);

			translate(pts_right_rotated, translation, pts_right_translated);
			pointsOnBlack(image, pts_left, pts_right_translated, "translatedpoints", j);

			fprintf(pFile, "%d, %f\n", j, ProcrustesDistance(pts_left, pts_right_translated, fp, j));

			sprintf(filename, "videocaps/frame%d.jpg", j);
			cvSaveImage(filename, image);
			CvMat* neutral_points_left = cvCreateMat(1,NUMLANDMARKS, CV_64FC1);
			cvWaitKey(1);
		}

		fclose(pFile);
		fclose(fp);
		cvReleaseImage(&image2);
	}
	
	else
	{
		IplImage* image = cvLoadImage(filename, -1);
		if(image == 0)
		{
			AAM_FormatMSG("ANNOT open image file %s\n", filename);
			AAM_ERROR(errmsg);
		}

		AAM_Shape Shape;
		bool flag = flag = model.InitShapeFromDetBox(Shape, facedet, image);
		if(flag == false) 
		{
			AAM_FormatMSG("The image doesn't contain any faces\n");
			AAM_ERROR(errmsg);
		}
		
		model.Fit(image, Shape, 50, true);
		model.Draw(image, Shape, 2);
		
		cvNamedWindow("Fitting");
		cvShowImage("Fitting", image);
		cvWaitKey(0);
		
		cvReleaseImage(&image);	
	}

	return 0;
}

/* gets the median x coordinate for the set of points */
float getMedian(CvMat* pts)
{
	float xvalues[(pts->cols) / 2];

	for (int i=0; i<(pts->cols / 2); i++) {
		xvalues[i] = cvGetReal1D(pts, (2*i));
	}

	int changes;
	float tmp;

	do {
		changes = 0;
		for (int j=0; j<(pts->cols / 2) - 1; j++) {
			if (xvalues[j] > xvalues[j+1]) {
				tmp = xvalues[j];
				xvalues[j] = xvalues[j+1];
    				xvalues[j+1] = tmp;
    				changes++;
			}
		}
	}
	while (changes);

	float median = (xvalues[(pts->cols / 2)/2] + xvalues[((pts->cols / 2)/2) - 1])/2;

	return median;
}

/* splits the set of points into two subsets, using the median x coordinate */
void splitPoints(CvMat* pts_left, CvMat* pts_right, CvMat* pts, float median)
{
	int left_cnt=0, right_cnt=0;

	for (int i=0; i<(pts->cols / 2); i++) {
		if (cvGetReal1D(pts, (2*i)) < median && left_cnt < pts_left->rows) {
			cvSetReal2D(pts_left, left_cnt, 0, cvGetReal1D(pts, (2*i)));
			cvSetReal2D(pts_left, left_cnt, 1, cvGetReal1D(pts, (2*i) + 1));
			left_cnt++;
		}
		else {
			if (right_cnt < pts_right->rows) {
				cvSetReal2D(pts_right, right_cnt, 0, cvGetReal1D(pts, (2*i)));
				cvSetReal2D(pts_right, right_cnt, 1, cvGetReal1D(pts, (2*i) + 1));
				right_cnt++;
			}
		}
	}
}

/* puts pts_left and pts_right on the image and shows the result */
void pointsOnImage(IplImage* image, CvMat* pts_left, CvMat* pts_right, int frameno)
{
	int x, y, x_, y_;
	IplImage* image2 = cvCreateImage(cvGetSize(image), image->depth, image->nChannels);
	cvCopy(image, image2);

	for (int i=0; i<pts_left->rows; i++) {
		x = floor(cvGetReal2D(pts_left, i, 0)+0.5);
		y = floor(cvGetReal2D(pts_left, i, 1)+0.5);
		x_ = floor(cvGetReal2D(pts_right, i, 0)+0.5);
		y_ = floor(cvGetReal2D(pts_right, i, 1)+0.5);
		for (int j=0;j<5;j++) {
			unsigned char* row1 = &CV_IMAGE_ELEM(image2, unsigned char, y - 2 + j, 0);
			unsigned char* row2 = &CV_IMAGE_ELEM(image2, unsigned char, y_ - 2 + j, 0);
			for (int k=0; k<5; k++) {
				if ((x-2+k) < image2->width && (x-2+k) >= 0) {
				  row1[(3*(x-2+k))+1] = 255;
				}
				if ((x_-2+k) < image2->width && (x_-2+k) >= 0) {
				  row2[(3*(x_-2+k))] = 255;
				  row2[(3*(x_-2+k))+1] = 255;
				  row2[(3*(x_-2+k))+2] = 100;
				}
			}
		}
	}

	cvShowImage("ptsOnImage", image2);
	cvWaitKey(1);

	char filename[100];

	if (inFrames_to_save(frameno)) {
		sprintf(filename, "ptsonimage%d.jpg", frameno);
		cvSaveImage(filename, image2);
	}

	cvReleaseImage(&image2);
}

/* prints out a matrix of points */
void printPts(CvMat* A)
{
	int rows = A->rows;
	for (int i=0; i<A->rows; ++i) {
		printf("Row %d: %f %f\n", i, cvGetReal2D(A, i, 0), cvGetReal2D(A, i, 1));
	}
	printf("\n");
}

/* prints an OpenCV matrix */
void PrintMat(CvMat *A)
{
	int i, j;
	for (i = 0; i < A->rows; i++)
	{
		printf("\n");
		switch (CV_MAT_DEPTH(A->type))
		{
			case CV_32F:
			case CV_64F:
			for (j = 0; j < A->cols; j++)
				printf ("%8.3f ", (float)cvGetReal2D(A, i, j));
				break;
			case CV_8U:
			case CV_16U:
			for(j = 0; j < A->cols; j++)
				printf ("%6d",(int)cvGetReal2D(A, i, j));
				break;
			default:
		break;
		}
	}
	printf("\n");
}

/* puts pts_left and pts_right on a black background and shows the result */
void pointsOnBlack(IplImage* image, CvMat* pts_left, CvMat* pts_right, char* windowname, int frameno)
{
	int x, y, x_, y_;
	IplImage* image2 = cvCreateImage(cvSize((image->width)*2, image->height), image->depth, image->nChannels);
	cvZero(image2);

	cvLine(image2, cvPoint(640,1), cvPoint(640,479), CV_RGB(255, 255, 255), 1, 8, 0);

	for (int i=0; i<pts_left->rows; i++) {
		x = floor(cvGetReal2D(pts_left, i, 0)+0.5+640);
		y = floor(cvGetReal2D(pts_left, i, 1)+0.5);
		x_ = floor(cvGetReal2D(pts_right, i, 0)+0.5+640);
		y_ = floor(cvGetReal2D(pts_right, i, 1)+0.5);
		for (int j=0;j<5;j++) {
			unsigned char* row1 = &CV_IMAGE_ELEM(image2, unsigned char, y - 2 + j, 0);
			unsigned char* row2 = &CV_IMAGE_ELEM(image2, unsigned char, y_ - 2 + j, 0);
			for (int k=0; k<5; k++) {
				if ((x-2+k) < image2->width && (x-2+k) >= 0) {
				  row1[(3*(x-2+k))+1] = 255;
				}
				if ((x_-2+k) < image2->width && (x_-2+k) >= 0) {
				  row2[(3*(x_-2+k))] = 255;
				  row2[(3*(x_-2+k))+1] = 255;
				  row2[(3*(x_-2+k))+2] = 100;
				}
			}
		}
	}

	cvShowImage(windowname, image2);
	cvWaitKey(1);

	char filename[100];

	if (inFrames_to_save(frameno)) {
		sprintf(filename, "%s%d.jpg", windowname, frameno);
		cvSaveImage(filename, image2);
	}

	cvReleaseImage(&image2);
}

/* performs Procrustes analysis on A and B to find the transformation matrix, T, and the translation vector */
void doProcrustes(CvMat* T, CvMat* translation, CvMat* A, CvMat* B)
{
	CvMat* ptssrc = cvCreateMat(A->rows, 2, CV_64FC1);
	CvMat* ptsdst = cvCreateMat(B->rows, 2, CV_64FC1);

	cvCopy(A, ptssrc);
	cvCopy(B, ptsdst);

	CvMat* srcmean = cvCreateMat(1, 2, CV_64FC1);
	CvMat* dstmean = cvCreateMat(1, 2, CV_64FC1);
	GetMean(ptssrc, srcmean);
	GetMean(ptsdst, dstmean);

	translate_to_origin(ptssrc);
	translate_to_origin(ptsdst);

	normalise(ptssrc);
	normalise(ptsdst);

	CvMat* srctrans = cvCreateMat(2, ptssrc->rows, CV_64FC1);
	cvTranspose(ptssrc, srctrans);

	CvMat* product = cvCreateMat(2, 2, CV_64FC1);
	cvGEMM(srctrans, ptsdst, 1, NULL, 0, product, 0);

	CvMat* L = cvCreateMat(2, 2, CV_64FC1);
	CvMat* D = cvCreateMat(2, 2, CV_64FC1);
	CvMat* M = cvCreateMat(2, 2, CV_64FC1);

	cvSVD(product, D, L, M);

	CvMat* Ltrans = cvCreateMat(2, 2, CV_64FC1);
	cvTranspose(L, Ltrans);
	cvGEMM(M, Ltrans, 1, NULL, 0, T, 0);

	CvMat* TproductmeanY = cvCreateMat(1, 2, CV_64FC1);
	cvGEMM(dstmean, T, 1, NULL, 0, TproductmeanY, 0);

	cvSub(srcmean, TproductmeanY, translation);
}

/* finds the centroid of a set of points */
void GetMean(CvMat* src, CvMat* dst)
{
	float sumx = 0, sumy = 0, avx = 0, avy = 0;

	for (int i = 0; i<src->rows; i++) {
	  sumx = sumx + cvGetReal2D(src, i, 0);
	  sumy = sumy + cvGetReal2D(src, i, 1);
	}

	avx = sumx / (src->rows);
	avy = sumy / (src->rows);

	cvSetReal1D(dst, 0, avx);
	cvSetReal1D(dst, 1, avy);
}

/* translates a set of points to the origin */
void translate_to_origin(CvMat* pts)
{
	float sumx = 0, sumy = 0, avx = 0, avy = 0;

	for (int i = 0; i<pts->rows; i++) {
	  sumx = sumx + cvGetReal2D(pts, i, 0);
	  sumy = sumy + cvGetReal2D(pts, i, 1);
	}

	avx = sumx / (pts->rows);
	avy = sumy / (pts->rows);

	for (int i = 0; i<pts->rows; i++) {
	  cvSetReal2D(pts, i, 0, cvGetReal2D(pts, i, 0) - avx);
	  cvSetReal2D(pts, i, 1, cvGetReal2D(pts, i, 1) - avy);
	}
}

/* normalises a set of points */
void normalise(CvMat* pts)
{
	float sumsq = 0;

	for (int i = 0; i<pts->rows; i++) {
	  sumsq = sumsq + (cvGetReal2D(pts, i, 0) * cvGetReal2D(pts, i, 0));
	  sumsq = sumsq + (cvGetReal2D(pts, i, 1) * cvGetReal2D(pts, i, 1));
	}

	sumsq = sqrt(sumsq);

	for (int i = 0; i<pts->rows; i++) {
	  cvSetReal2D(pts, i, 0, cvGetReal2D(pts, i, 0) / sumsq);
	  cvSetReal2D(pts, i, 1, cvGetReal2D(pts, i, 1) / sumsq);
	}
}

/* translates a set of points by the translation vector */
void translate(CvMat* src, CvMat* translation, CvMat* dst)
{
  for (int i=0; i<src->rows; i++) {
	  cvSetReal2D(dst, i, 0, cvGetReal2D(src, i, 0) + cvGetReal1D(translation, 0));
	  cvSetReal2D(dst, i, 1, cvGetReal2D(src, i, 1) + cvGetReal1D(translation, 1));
  }
}

/* calculates the Procrustes distance between two sets of points */
float ProcrustesDistance(CvMat* one, CvMat* two, FILE* f, int frameno)
{
	float sum = 0;
	for (int i=0; i<one->rows; i++) {
		sum = sum + (pow(cvGetReal2D(one, i, 0) - cvGetReal2D(two, i, 0),2));
		sum = sum + (pow(cvGetReal2D(one, i, 1) - cvGetReal2D(two, i, 1),2));
		fprintf(f, "%d %f %f\n", frameno, abs(cvGetReal2D(one, i, 0) - cvGetReal2D(two, i, 0)),	abs(cvGetReal2D(one, i, 1) - cvGetReal2D(two, i, 1)));
	}
	sum = sqrt(sum);
	return sum;
}

/* self-explanatory */
bool inFrames_to_save(int a)
{
	for (int i=0; i<FRAMESTOPRINT; i++) {
		if (a==frames_to_save[i]) {
			return 1;
		}
	}

	return 0;
}

/* overlays a set of points on an image */
void overlayPoints(IplImage* dst, IplImage* image, CvMat* points, int colour)
{
  int x, y;
  cvCopy(image, dst);

  for (int i=0; i<points->cols; i=i+2) {
    y = floor(cvGetReal1D(points, i+1)+0.5);
    x = floor(cvGetReal1D(points, i)+0.5);
    for (int j=0;j<5;j++) {
    	unsigned char* row1 = &CV_IMAGE_ELEM(dst, unsigned char, y-2+j, 0);
    	for (int k=0; k<5; k++) {
    		row1[(3*(x-2+k))+colour] = 255;
    	}
    }
  }
}
\end{lstlisting}
\end{document}